\documentclass{article}
\usepackage{iclr2027_conference,times}

\usepackage{amsmath,amsfonts,bm}

\def\eqref#1{equation~\ref{#1}}

\def\1{\bm{1}}

\DeclareMathAlphabet{\mathsfit}{\encodingdefault}{\sfdefault}{m}{sl}
\SetMathAlphabet{\mathsfit}{bold}{\encodingdefault}{\sfdefault}{bx}{n}

\usepackage[utf8]{inputenc}
\usepackage[T1]{fontenc}
\usepackage{hyperref}
\usepackage{url}
\usepackage{booktabs}
\usepackage{amsfonts}
\usepackage{amsmath}
\usepackage{nicefrac}
\usepackage{microtype}
\usepackage{xcolor}
\usepackage{graphicx}
\usepackage{subcaption}
\usepackage{multirow}
\usepackage{placeins}

\title{SPIRE: Shared-Private Representation Learning for Continuous Multi-Region LFP under Perturbation}

\author{
Rahil Soroushmojdehi$^{1}$, Sina Javadzadeh$^{2}$, Mehrnaz Asadi$^{1}$, Terence D.~Sanger$^{1,3}$\thanks{Corresponding author: \texttt{terry@sangerlab.net}}\\[4pt]
$^{1}$Department of Electrical Engineering and Computer Science, University of California, Irvine, CA, USA\\
$^{2}$Department of Biomedical Engineering, University of California, Irvine, CA, USA\\
$^{3}$Department of Neurology, Children’s Health Orange County, Orange, CA, USA
}

\iclrfinalcopy 
\begin{document}
\maketitle

\pagestyle{fancy}
\fancyhf{} 
\lhead{\textit{Preprint. Submitted to ICLR 2027.}}
\rhead{} 
\cfoot{\thepage} 

\begin{abstract}
Learning shared and region-specific representations from continuous multi-region neural signals is challenging when observations are nonlinear, autocorrelated, spectrally structured, and temporally misaligned. These challenges are particularly pronounced in local field potentials (LFPs), where separating activity that is coordinated across regions from activity specific to individual regions is important for studying distributed neural systems and their response to intervention. We introduce \textbf{SPIRE} (Shared-Private Inter-Regional Encoder), a nonlinear, time-resolved framework for learning shared and private representations from continuous multi-region LFP. SPIRE combines region-specific sequence encoders with constrained temporal correspondence, shared-only reconstruction, and variance regularization for shared/private separation. On LFP-oriented synthetic benchmarks with known ground-truth factors, SPIRE preferentially assigns shared and private information to the intended latent spaces under nonlinear observation mappings, time-varying delays, and spectral confounding. On held-out human GPi-STN LFP, nonlinear prediction reveals strong transfer of shared representations across regions, whereas cross-region prediction from private representations remains at a structured temporal null, supporting functional rather than strict statistical separation. We then freeze representations learned from off-stimulation recordings and apply them to intracranial deep brain stimulation (DBS) data. Stimulation induces larger changes and carries more condition information in shared than private representations. A matched CTAE analysis reproduces the qualitative shared-over-private effect while retaining substantially less stimulation information. Together, these results establish SPIRE as a shared/private representation-learning framework for continuous multi-region LFP and show that DBS-related changes are preferentially expressed in representations shared across the recorded regions.
\end{abstract}

\section{Introduction}
Learning low-dimensional representations from multiple neural populations requires separating structure that is shared across regions from structure that is specific to each region. This becomes difficult for continuous neural signals, where observations can be strongly autocorrelated, nonlinear, spectrally structured, and temporally misaligned. In such settings, high reconstruction quality alone does not establish that learned ``shared'' and ``private'' representations actually capture transferable and region-specific information.

Local field potentials (LFPs) provide an important instance of this problem. LFPs contain continuous oscillatory and broadband activity, strong temporal autocorrelation, nonstationarity, and substantial $1/f$ structure~\citep{bedard2006does,he2014scale,dupre2017non}. Related processes may also appear differently across recording sites because of region-specific nonlinear mixing and temporal offsets. These properties can confound shared/private decomposition, for example by assigning dominant low-frequency or high-variance activity to the shared space even when it is not truly cross-regional.

A reliable decomposition is particularly useful for studying deep brain stimulation (DBS). DBS acts on distributed basal ganglia-thalamo-cortical circuits~\citep{benabid2003deep,lozano2019deep, larsh2021deep,mcintyre2010network,horn2020opportunities,yang2021modelling},  yet many intracranial analyses characterize stimulation through local quantities such as spectral power or evoked responses~\citep{dale2022evoked,milosevic2018physiological,schmidt2020evoked,soroushmojdehi2025frequency}. Complementary work has investigated inter-regional LFP signal transmission and its relationship to structural connectivity~\citep{kasiri2025local}. During ongoing stimulation, population-level spike measurements can be difficult to obtain reliably because stimulation artifacts, amplifier blanking or saturation, and disrupted spike detection and sorting can obscure spiking activity. Multi-region LFP provides an opportunity to examine whether stimulation-related changes are expressed primarily in region-specific activity or in neural representations shared across regions.

Existing latent-variable methods address parts of this problem. Linear approaches such as CCA and DLAG can identify shared structure across neural populations~\citep{bach2005probabilistic,gokcen2022disentangling}, but their linear or linear-Gaussian assumptions can be restrictive for continuous LFP. More recent nonlinear approaches, particularly CTAE, learn time-resolved shared and private neural representations using coupled autoencoders, latent alignment, and cross-region reconstruction~\citep{sristi2025coupled}. However, it remains unclear whether shared/private representations retain their intended functional roles under the nonlinear mixing, temporal variability, and spectral confounding characteristic of continuous multi-region LFP. Evaluating only reconstruction or intended-factor recovery is insufficient to establish that information has been preferentially assigned to the correct latent space.

We introduce \textbf{SPIRE} (Shared-Private Inter-Regional Encoder), a nonlinear, time-resolved framework for shared/private representation learning in continuous multi-region LFP. SPIRE uses region-specific sequence encoders and decoders to produce shared and private latent trajectories, with constrained temporal correspondence and shared-only reconstruction encouraging transferable cross-regional structure, and cross-covariance and variance regularization promoting functional separation between the two spaces.

We evaluate SPIRE on LFP-oriented synthetic benchmarks spanning linear, nonlinear, temporally misaligned, and spectrally confounded regimes, using complete intended-versus-crossed recovery to test whether shared and private information is preferentially assigned to the correct latent spaces. We then validate this distinction on held-out human GPi-STN LFP using nonlinear cross-region prediction against a structured temporal null. Finally, we freeze representations learned from off-stimulation recordings and use them to test whether DBS preferentially reorganizes shared or private activity, with a matched CTAE analysis assessing whether the resulting stimulation effects generalize across shared/private models.

Our main contributions are:
\begin{enumerate}
    \item \textbf{Shared/private representation learning for continuous multi-region LFP.}
    We introduce SPIRE, combining nonlinear region-specific sequence models, constrained temporal correspondence, shared-only reconstruction, and variance constraints for shared/private separation.

    \item \textbf{Direct validation of functional shared/private separation.}
    We evaluate complete intended-versus-crossed recovery under synthetic ground truth and nonlinear cross-region predictability in held-out human LFP, showing that shared representations preferentially retain transferable structure while private representations retain predominantly region-specific structure.

    \item \textbf{Application to distributed effects of human DBS.}
    DBS produces larger representational changes and more strongly encodes stimulation condition in shared than private latents. A matched CTAE analysis reproduces the direction of this result, while SPIRE retains substantially more stimulation information.
\end{enumerate}

\section{Related work}
\paragraph{Multi-region latent-variable models.}
Classical latent models such as GPFA~\citep{yu2008gaussian}, CCA~\citep{bach2005probabilistic}, and DLAG~\citep{gokcen2022disentangling} provide principled low-dimensional descriptions of neural population activity, with DLAG explicitly separating across-population and within-population latent structure. Their linear or linear-Gaussian observation assumptions are well matched to some neural population datasets but can be restrictive for spectrally rich LFP. Nonlinear and dynamical approaches including LFADS~\citep{pandarinath2018inferring}, MARBLE~\citep{gosztolai2025marble}, and MRDS-IR~\citep{dowlingnonlinear} address complementary questions such as latent trajectory modeling or inter-region dynamics. SPIRE instead targets descriptive time-resolved shared/private representation learning; it does not estimate a state-transition law, directed communication, unobserved inputs, or future responses to unseen perturbations. Other related approaches include MR-SDS~\citep{karniol2024modeling}, and MR-LFADS~\citep{liu2025accurateMRLFADS}, which target switching multi-region dynamics, and inter-regional communication with latent inputs, respectively. Recent studies have explored complementary aspects of multi-region LFP, including functional electrode embeddings for cross-subject aggregation~\citep{javadzadeh2025functional} and BACE for behavior-dependent directed connectivity estimation~\citep{asadi2025bace}. In contrast, SPIRE focuses on disentangling shared and region-specific latent dynamics from continuous multi-region LFP.

\paragraph{Shared/private nonlinear representation learning.}
Shared/private factorization is common in multimodal representation learning~\citep{yishared,shi2019variational,lee2021private}. CTAE~\citep{sristi2025coupled} is the closest neural baseline because it learns time-resolved shared and region-specific latent trajectories with transformer-based coupling, cross-reconstruction, alignment, and orthogonality. SPIRE differs in using region-specific recurrent encoders/decoders; near-identity temporal alignment followed by controlled linear mappings; variance guards for both branches; shared-self and aligned cross-shared reconstruction; and staged private activation/freezing. The contribution is not any single module in isolation, but the combined formulation and its validation under nonlinear, delayed, and spectrally confounded LFP conditions.

\paragraph{LFP and perturbation analysis.}
LFP reflects complementary aspects of neural population activity and poses signal-processing challenges distinct from spikes~\citep{bedard2006does,he2014scale,erturk2025cross}. During DBS, simultaneous multi-region LFP is particularly useful for examining distributed network effects despite stimulation-related recording constraints~\citep{alarie2022artifact}. Recent work further highlights that spiking and LFP signals capture complementary aspects of neural activity and require distinct modeling approaches~\citep{erturk2025cross}. Our perturbation analysis therefore learns the representation without stimulation recordings or labels, freezes it, and measures off-to-on changes in the resulting coordinate system. This design avoids circularly shaping the representation toward the downstream stimulation labels.

\section{SPIRE for continuous multi-region LFP}
\label{sec:method}
\subsection{Model overview}
\begin{figure}[t]
\centering
\includegraphics[width=0.62\linewidth,clip=true,trim=0in 0in 2.5in 0in]{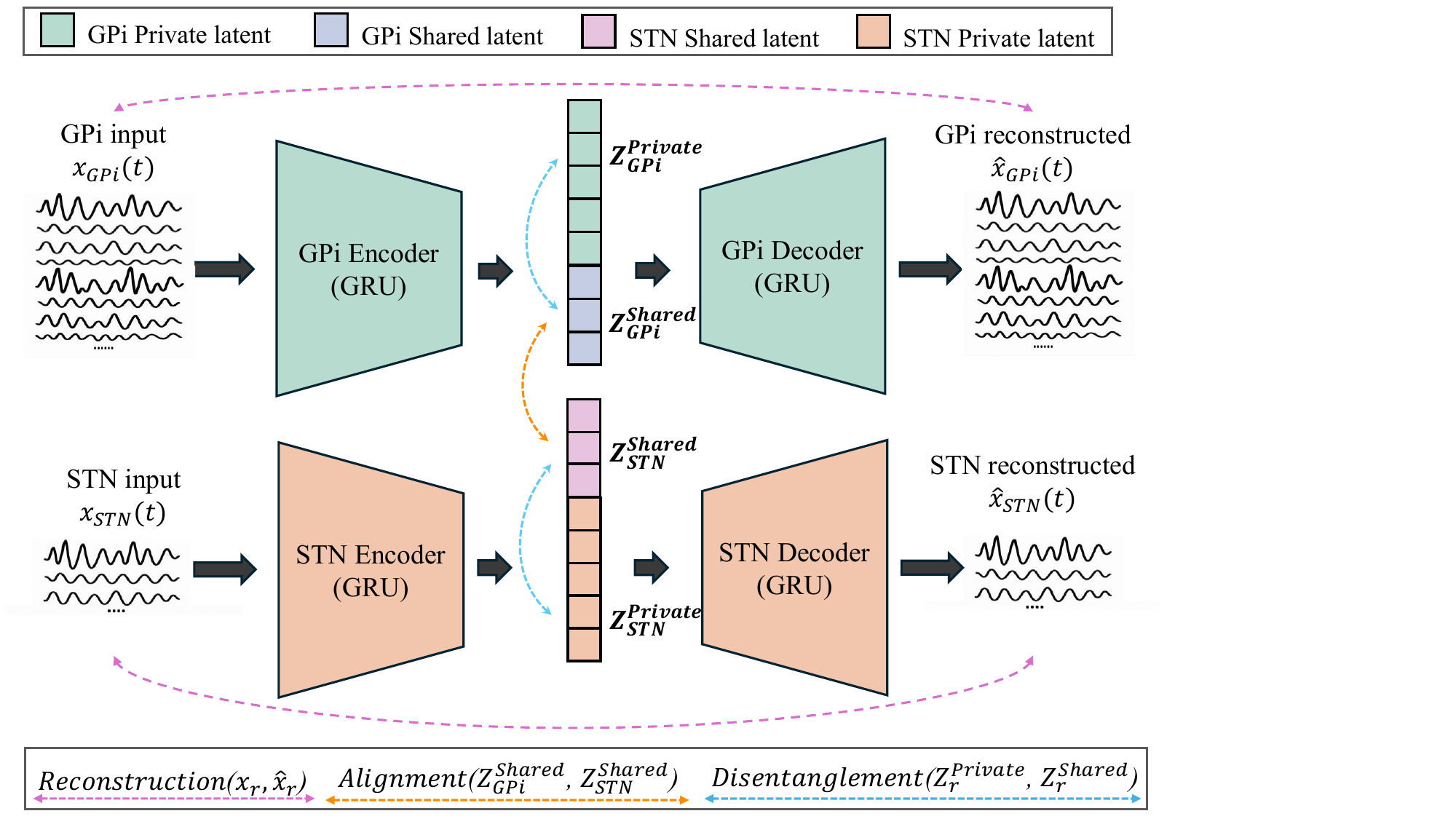}
\caption{SPIRE separates each region into time-resolved shared and private latent trajectories. Shared trajectories are aligned across regions through constrained temporal and linear mappings, while shared/private separation and reconstruction objectives preserve complementary information.}
\label{fig:schematic}
\end{figure}

We introduce \textbf{SPIRE} (Shared-Private Inter-Regional Encoder), a
nonlinear, time-resolved framework that learns shared (cross-regional) and
private (region-specific) latent representations from multi-region neural
recordings. Figure ~\ref{fig:schematic} shows a schematic of SPIRE for a two-region example (GPi and STN).

For each region $r \in \mathcal{R}$, SPIRE instantiates an encoder–decoder pair. Given multichannel input $x^{(r)} \in \mathbb{R}^{B \times T \times C_r}$ (batch $B$, length $T$, $C_r$ channels), a recurrent encoder produces hidden states $h^{(r)}$, which are linearly projected into shared and private latent sequences ($d_{\text{sh}},d_{\text{pr}}$ are shared and private latent dimensionalities):
\begin{align}
z^{(r)}_{\text{sh}} &= W^{(r)}_{\text{sh}} h^{(r)} \in \mathbb{R}^{B \times T \times d_{\text{sh}}}, \qquad
z^{(r)}_{\text{pr}} = W^{(r)}_{\text{pr}} h^{(r)} \in \mathbb{R}^{B \times T \times d_{\text{pr}}},
\end{align}

Each region is reconstructed from its own latents:
\begin{align}
\hat{x}^{(r)} = f^{(r)}_{\text{dec}}\!\left([\,z^{(r)}_{\text{sh}},\, z^{(r)}_{\text{pr}}\,]\right) 
\in \mathbb{R}^{B \times T \times C_r}.
\end{align}

The terms ``shared'' and ``private'' are operational: shared structure should preferentially transfer across the modeled regions, whereas private structure should preferentially remain region specific. They are not assumed to be statistically independent biological sources.

\paragraph{Constrained temporal correspondence.}
Shared trajectories from region $s$ are mapped into region $r$ as
\begin{equation}
 \tilde z^{(s\rightarrow r)}_{\mathrm{sh}}=M^{(s\rightarrow r)}\,\mathrm{ConvAlign}^{(s\rightarrow r)}(z^{(s)}_{\mathrm{sh}}),
\end{equation}
where the depthwise temporal convolution and linear mapper are initialized near identity and regularized to remain close to identity. This permits local temporal/subspace adjustment without allowing an unconstrained transformation to define the shared space.

\subsection{Training objective}
The objective has three conceptual families,
\begin{equation}
\mathcal{L}=\lambda_{\mathrm{rec}}\mathcal{L}_{\mathrm{rec}}^{\mathrm{family}}+
\lambda_{\mathrm{align}}\mathcal{L}_{\mathrm{align}}^{\mathrm{family}}+
\lambda_{\mathrm{dis}}\mathcal{L}_{\mathrm{dis}}^{\mathrm{family}}.
\end{equation}
The reconstruction family contains full reconstruction, shared-self reconstruction, and aligned cross-region shared reconstruction. The alignment family uses variance–invariance–covariance regularization objective (VICReg, \citealp{bardes2022vicreg}) between native and aligned shared trajectories together with near-identity mapper/ConvAlign regularization. Explicitly,
\begin{equation}
\mathcal{L}_{\mathrm{VICReg}}(A,B)=\lambda_{\mathrm{inv}}\mathcal{L}_{\mathrm{inv}}+\lambda_{\mathrm{var}}\mathcal{L}_{\mathrm{var}}+\lambda_{\mathrm{cov}}\mathcal{L}_{\mathrm{cov}},
\end{equation}
where invariance aligns corresponding trajectories, the variance term prevents collapse, and the covariance term discourages redundant dimensions. The disentanglement family penalizes standardized shared-private cross-covariance and includes shared/private variance guards. Full equations, coefficient definitions, and optimization schedules are provided
in Appendix~\ref{app:spire_losses}. The objective contains a fixed set of coefficient types: region-wise terms are averaged across regions and pairwise terms across directed region pairs, so adding a region adds modules and computation but not pair-specific loss-weight types.

\subsection{Implementation and evaluation protocol}
Encoders/decoders are GRU sequence models trained with Adam, gradient clipping, learning-rate scheduling, and early stopping. Short lag-stacked inputs provide local within-region context, while ConvAlign acts on encoded shared trajectories across regions; all compared real-data models receive identical lag augmentation. Detailed preprocessing, schedules, loss ablations, and a no-lag ablation are given in Appendix~\ref{app:spire_impl}, Appendix~\ref{app:ablation} and Appendix~\ref{app:robustness}.

For human DBS analyses, SPIRE is trained and selected per hemisphere using off-stimulation data only, then frozen. Stimulation recordings are used only for downstream displacement and decoding analyses. Appendix~\ref{app:train_test_on} verifies that the frozen representation remains reconstructive and functionally organized under stimulation.

SPIRE is implemented and evaluated in PyTorch~\citep{paszke2019pytorch}, statistical analyses were performed in R~\citep{RCoreTeam2025} using RStudio~\citep{PositTeam2025}.

\section{Controlled LFP-oriented validation}
\label{synth}

A key advantage of synthetic data is that the true shared and private latent
processes are known. This allows us to test directly both whether a model
recovers the underlying factors and whether it assigns them to the intended
shared or private representation. We generated four regimes with three shared
and three private latent factors per region. D0 is an approximately linear
positive control; D1 adds nonlinear region-specific warping and nonlinear
observation mixing; D2 additionally introduces time-varying inter-region delay;
and D3 deliberately makes strong private structure low-frequency while shared
structure occupies higher-frequency bands, creating an LFP-like spectral
confound. The generator additionally includes colored noise, autocorrelation, common-mode
structure, and bipolar rereferencing; full generation details and parameter
settings are provided in Appendix~\ref{app:datagen}. We compare SPIRE with
DLAG~\citep{gokcen2022disentangling} and CTAE~\citep{sristi2025coupled}
using four matched seeds (implementation details mentioned in Appendix~\ref{app:synth_implementation},~\ref{app:dlag_fit} and~\ref{app:ctae_fit_synth}).

\paragraph{Ground-truth recovery.}
Let $S_{\mathrm{GT}}$ denote the ground-truth shared latent factors used to
generate activity across regions, and let $P_{\mathrm{GT}}$ denote the
ground-truth private factors specific to the evaluated region. Similarly,
$Z_{\mathrm{sh}}$ and $Z_{\mathrm{pr}}$ denote the corresponding learned
shared and private representations. We measure nonlinear recovery using Extra
Trees regression~\citep{geurts2006extremely} with five-fold trial-wise cross-validation and report the
held-out coefficient of determination, $R^2$.

We first evaluate \emph{intended recovery}: shared recovery,
$R^2(Z_{\mathrm{sh}}\rightarrow S_{\mathrm{GT}})$, and private recovery,
$R^2(Z_{\mathrm{pr}}\rightarrow P_{\mathrm{GT}})$. Table~\ref{tab:synth_recovery}
shows that DLAG performs best in the linear D0 positive control, whereas SPIRE
achieves the strongest intended recovery in each nonlinear regime.

\begin{figure}[t]
  \centering
  \begin{subfigure}[b]{0.48\textwidth}
    \centering
    \includegraphics[width=\textwidth]{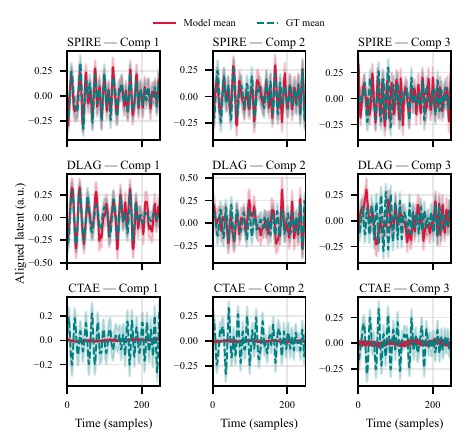}
    \caption{Shared latent recovery.}
  \end{subfigure}
  \hfill
  \begin{subfigure}[b]{0.48\textwidth}
    \centering
    \includegraphics[width=\textwidth]{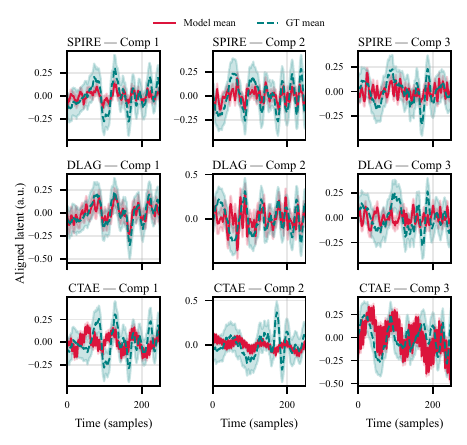}
    \caption{Private latent recovery.}
  \end{subfigure}
  \caption{
  Representative latent recovery in the challenging D3 synthetic regime.
  Recovered trajectories (solid red) are compared with ground-truth trajectories
  (dashed teal) after CCA-based alignment for visualization.
  D3 combines nonlinear observation mappings with strong low-frequency private
  structure and higher-frequency shared factors, producing substantial spectral
  confounding. Shaded regions denote SEM across trials. Quantitative nonlinear
  recovery is reported in Table~\ref{tab:synth_recovery}, and branch specificity
  in Table~\ref{tab:synth_specificity}.
  }
  \label{fig:synth_alignment_example}
\end{figure}

Figure~\ref{fig:synth_alignment_example} provides a representative visualization
from D3, the most spectrally confounded regime. SPIRE tracks both shared and
private ground-truth trajectories closely, whereas the baseline representations
show larger deviations.

\begin{table}[t]
\caption{
Nonlinear recovery of known ground-truth factors.
Each cell reports intended shared/private recovery,
$R^2(Z_{\mathrm{sh}}\!\rightarrow\!S_{\mathrm{GT}})
\,/\,
R^2(Z_{\mathrm{pr}}\!\rightarrow\!P_{\mathrm{GT}})$,
averaged across four matched seeds. Higher values indicate stronger recovery
of the corresponding ground-truth factors. Full means and standard deviations
are reported in Appendix~\ref{app:nonlinear_matrix}.
}
\label{tab:synth_recovery}
\centering
\small
\begin{tabular}{lccc}
\toprule
Regime & SPIRE & DLAG & CTAE \\
\midrule
D0 linear
& 0.72 / 0.64
& \textbf{0.89 / 0.83}
& 0.29 / 0.31 \\

D1 nonlinear
& \textbf{0.76 / 0.60}
& 0.64 / 0.48
& 0.14 / 0.17 \\

D2 + delay
& \textbf{0.75 / 0.58}
& 0.57 / 0.48
& 0.12 / 0.16 \\

D3 spectral confound
& \textbf{0.54 / 0.55}
& 0.42 / 0.46
& 0.17 / 0.35 \\
\bottomrule
\end{tabular}
\end{table}

\paragraph{Intended-versus-crossed recovery.}
High intended recovery alone is not sufficient to establish a meaningful
shared/private decomposition: for example, a learned shared representation
could recover the true shared factors while also retaining substantial private
information. We therefore evaluate the complete $2\times2$ recovery matrix.

We define
\begin{align}
SS &= R^2(Z_{\mathrm{sh}}\rightarrow S_{\mathrm{GT}}),
&
SP &= R^2(Z_{\mathrm{sh}}\rightarrow P_{\mathrm{GT}}), \\
PP &= R^2(Z_{\mathrm{pr}}\rightarrow P_{\mathrm{GT}}),
&
PS &= R^2(Z_{\mathrm{pr}}\rightarrow S_{\mathrm{GT}}).
\end{align}
Here, $SS$ and $PP$ measure recovery of the factors intended for each branch,
whereas $SP$ and $PS$ quantify crossed recovery: private information recoverable
from the learned shared space and shared information recoverable from the learned
private space, respectively.

We summarize branch specificity as
\begin{equation}
\Delta_{\mathrm{sh}} = SS-SP,
\qquad
\Delta_{\mathrm{pr}} = PP-PS.
\end{equation}
Positive $\Delta_{\mathrm{sh}}$ therefore means that the learned shared
representation predicts true shared factors better than true private factors,
while positive $\Delta_{\mathrm{pr}}$ means that the learned private
representation preferentially predicts true private factors.
Table~\ref{tab:synth_specificity} reports these specificity margins; the full
$2\times2$ recovery matrices with standard deviations are given in
Appendix~\ref{app:nonlinear_matrix}.

\begin{table}[t]
\caption{
Nonlinear shared/private specificity.
Each cell reports
$\Delta_{\mathrm{sh}}/\Delta_{\mathrm{pr}}$
(mean across four matched seeds), where
$\Delta_{\mathrm{sh}}=SS-SP$ and
$\Delta_{\mathrm{pr}}=PP-PS$.
Positive values indicate preferential assignment of information to the intended
latent branch.
}
\label{tab:synth_specificity}
\centering
\small
\begin{tabular}{lccc}
\toprule
Regime & SPIRE & DLAG & CTAE \\
\midrule
D0 linear
& 0.48 / 0.40
& \textbf{0.81 / 0.68}
& 0.07 / -0.01 \\

D1 nonlinear
& \textbf{0.44 / 0.39}
& 0.43 / 0.20
& 0.02 / -0.01 \\

D2 + delay
& \textbf{0.42 / 0.38}
& 0.37 / 0.16
& 0.04 / -0.01 \\

D3 spectral confound
& \textbf{0.42 / 0.27}
& 0.15 / 0.13
& -0.06 / 0.14 \\
\bottomrule
\end{tabular}
\end{table}

The two analyses provide complementary evidence. D0 acts as a positive control: in the approximately linear regime, which is
closest to DLAG's linear-Gaussian assumptions, DLAG achieves both the
strongest ground-truth recovery and the strongest shared/private specificity.
In the nonlinear D1 and delayed D2 regimes, SPIRE achieves the strongest
intended recovery while maintaining positive specificity in both branches; its
private specificity is substantially larger than DLAG's in both settings. In
the challenging D3 regime, SPIRE achieves shared/private specificity margins of
$0.42/0.27$, compared with $0.15/0.13$ for DLAG and $-0.06/0.14$ for CTAE.
Because D3 deliberately places the shared factors at higher frequencies than
the dominant private factors, this result argues against the explanation that SPIRE obtains its decomposition simply by assigning slower LFP structure to the shared branch.
Remaining recovered trajectories of D0-D2 regimes, together
with complementary CCA-based recovery analyses, are shown in
Appendix~\ref{app:synth_cca_results}.

\section{Functional shared/private organization in human LFP}
\label{disentangle_real}
We analyzed intracranial LFP from ten pediatric dystonia patients with recordings from GPi, STN, and, in a subset, thalamus~\citep{liker2019pediatric,liker2024stereotactic,sanger2018case}. The principal analyses use paired GPi-STN recordings from 17 hemispheres; stimulation analyses use 17 GPi-stimulation hemispheres and 13 STN-stimulation hemispheres. Signals were band-limited to suppress stimulation artifacts, bipolar rereferenced, and segmented into 0.5-s windows. Models were fit separately by hemisphere because channel count, electrode placement, anatomy, and recording duration differ across subjects and hemispheres. Shared
and private latent dimensions were selected from a predefined grid using
reconstruction performance and variance partitioning criteria. Dataset details,
latent-dimension selection, and training details are provided in
Appendix~\ref{app:real_info}-\ref{app:human_training} and Appendix ~\ref{app:baseline_training}.

\subsection{Nonlinear cross-region predictability defines functional segregation}

True biological shared/private factors are unobserved in human data, so we use
held-out nonlinear predictability to test the operational role of each branch.
Extra Trees regressors were evaluated with five-fold cross-validation holding out
complete 0.5-s windows. A structured temporal null reassigns complete target
windows and circularly shifts trajectories, preserving marginal distributions
and temporal autocorrelation. We report excess $R^2$ relative to this null.

\begin{table}[t]
\caption{
Held-out nonlinear organization of human GPi-STN LFP.
Cross-region shared prediction is strongly above the structured temporal null,
whereas cross-region private prediction is approximately at the null.
Values are hemisphere-level mean excess $R^2$ with 95\% confidence intervals.
}
\label{tab:human_nonlinear}
\centering
\small
\begin{tabular}{lcc}
\toprule
Prediction category & Excess $R^2$ & 95\% CI \\
\midrule
Shared across regions & \textbf{0.636} & [0.507, 0.764] \\
Shared $\rightarrow$ private, within region & 0.504 & [0.376, 0.633] \\
Private $\rightarrow$ shared, within region & 0.312 & [0.183, 0.440] \\
Private across regions & \textbf{0.009} & [-0.120, 0.138] \\
\bottomrule
\end{tabular}
\end{table}

As shown in Table~\ref{tab:human_nonlinear}, shared representations transfer strongly across GPi and STN, with mean excess $R^2=0.636$, whereas
cross-region prediction from private representations remains near the structured
null ($R^2=0.009$, $p=0.443$). Within each region, shared and private
representations retain substantial nonlinear predictability in both directions,
indicating that the learned branches are not statistically independent.

Together with the synthetic ground-truth recovery and specificity results in
Tables~\ref{tab:synth_recovery} and~\ref{tab:synth_specificity}, these findings
support a narrower interpretation of \emph{functional segregation}: shared
representations preferentially capture information that transfers between
recorded regions, whereas private representations remain predominantly
region specific. Complementary human-data comparisons with CTAE using held-out reconstruction
and CCA-based latent organization are reported in
Appendix~\ref{app:human_secondary}.

\begin{figure*}[t]
\centering

\begin{subfigure}[t]{0.48\linewidth}
\centering
\includegraphics[width=\linewidth]{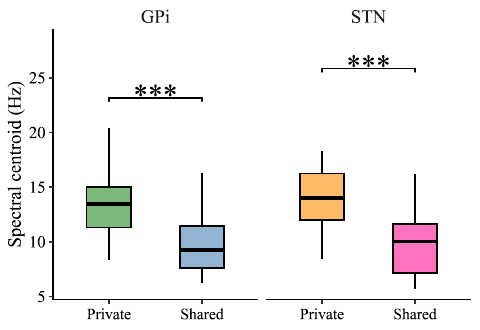}
\caption{Spectral organization of latent representations.}
\label{fig:spec_cent}
\end{subfigure}
\hfill
\begin{subfigure}[t]{0.48\linewidth}
\centering
\includegraphics[width=\linewidth,height=0.75\linewidth]{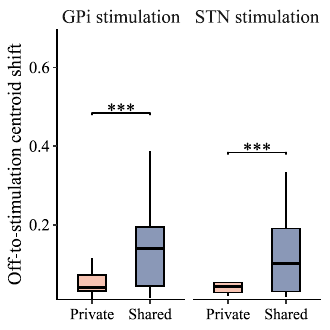}
\caption{Off-to-on latent displacement.}
\label{fig:cent_shift}
\end{subfigure}

\vspace{0.5em}

\begin{subfigure}[t]{0.48\linewidth}
\centering
\includegraphics[width=\linewidth]{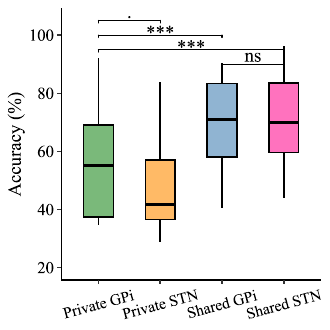}
\caption{GPi stimulation decoding.}
\label{fig:gpi_classify}
\end{subfigure}
\hfill
\begin{subfigure}[t]{0.48\linewidth}
\centering
\includegraphics[width=\linewidth]{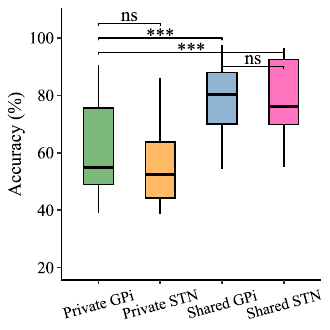}
\caption{STN stimulation decoding.}
\label{fig:stn_classify}
\end{subfigure}

\caption{
Functional organization and DBS-induced changes in SPIRE representations.
(a) Shared and private latent trajectories exhibit distinct spectral structure,
with shared latents showing lower spectral centroids in human GPi-STN LFP.
(b) DBS produces larger off-to-on displacement in shared than private
representations.
(c,d) Shared representations encode stimulation condition more strongly than
private representations under both GPi and STN stimulation.
Statistical significance:
$*:p<0.05$,
$**:p<0.01$,
$***:p<0.001$; ns, not significant.
}
\label{fig:human_latent_dbs}
\end{figure*}

\paragraph{Shared and private representations exhibit distinct spectral structure.}
Beyond cross-region transferability, the learned representations also differ in
their temporal organization. Shared latents exhibit significantly lower spectral
centroids than private latents in both GPi and STN
(Fig.~\ref{fig:spec_cent}), indicating that the latent structure that transfers across regions in these recordings is associated with slower spectral content than the
predominantly region-specific structure. Importantly, this separation is not
imposed by the model: in the D3 synthetic benchmark, shared ground-truth factors
were deliberately placed at higher frequencies than the dominant private factors,
yet SPIRE still preferentially recovered the correct shared and private factors.
Thus, the spectral difference observed in human LFP reflects structure in the
data rather than a generic low-frequency bias of the learned decomposition.

\section{DBS-related changes are preferentially expressed in the shared representation}
\label{net_reorg}
Having established the functional roles of the two latent spaces on held-out off-stimulation data, we freeze each selected model and analyze unseen stimulation recordings. Additional robustness analyses, including train-off/test-off/test-on
evaluation, are reported in Appendix~\ref{app:train_test_on}. We quantify off-to-on centroid displacement and decode stimulation frequency from each latent subset with Random Forest classifiers~\citep{breiman2001random}.

Shared latents exhibit significantly larger off-to-on displacement than private latents for both GPi and STN stimulation (Figure~\ref{fig:cent_shift}). For GPi stimulation, classifiers distinguish Off, 85, 185, and 250~Hz across 17 hemispheres; for STN stimulation, they distinguish Off, 85, and 185~Hz across 13 hemispheres. Shared latents provide consistently higher decoding accuracy than private latents across regions and stimulation sites (Figures~\ref{fig:gpi_classify} and~\ref{fig:stn_classify}). Thus stimulation-related information is preferentially expressed in the learned shared representation rather than only in the learned region-specific residual representation.

\subsection{Matched CTAE analysis}
To test whether the qualitative neuroscience conclusion depends on SPIRE, we repeated the GPi-stimulation decoding analysis with CTAE using the same 17 hemispheres, preprocessing, lag-augmented inputs, matched latent dimensionalities, stimulation tensors, and classifier procedure. CTAE independently reproduces shared $>$ private decoding: its mean shared accuracy is 38.4\% versus 31.0\% private, a 7.41-point difference ($F_{1,48}=44.53$, $p=2.34\times10^{-8}$). However, SPIRE retains substantially more stimulation information in both spaces (Table~\ref{tab:ctae_stim}, details in ~\ref{app:CTAE_onstim_additional}).

\begin{table}[t]
\caption{Matched GPi-stimulation decoding comparison. Accuracies are averaged across regions after first averaging stimulation-contact settings within hemisphere. Both SPIRE-CTAE differences have Holm-adjusted $p<0.0001$.}
\label{tab:ctae_stim}
\centering
\small
\begin{tabular}{lccc}
\toprule
Latent class & CTAE & SPIRE & Difference \\
\midrule
Private & 31.0\% & 53.6\% & +22.6 points \\
Shared & 38.4\% & 67.7\% & +29.3 points \\
\bottomrule
\end{tabular}
\end{table}

This agreement is useful: the shared-over-private stimulation effect is not unique to SPIRE. At the same time, SPIRE retains substantially more stimulation-related information under this decoding evaluation. The difference-of-differences between the models' shared-minus-private gaps is not significant ($p=0.076$), so we do not claim that SPIRE changes the qualitative conclusion; rather, it preserves considerably more stimulation information while recovering the intended shared/private roles more reliably in the LFP-oriented validation.

\subsection{Extension to three recorded regions.}
Although the primary analyses focus on GPi-STN recordings, we also instantiated
SPIRE jointly on GPi, STN, and thalamic LFP. The three-region model preserved
the expected latent organization, with shared representations showing the
strongest cross-region CCA similarity and private representations the weakest.
Shared latents from all three regions also encoded GPi stimulation frequency
more strongly than their corresponding private latents. These results demonstrate
practical extension beyond two regions, while not implying unrestricted
many-region scalability. Implementation details and full three-region analyses are reported in
Appendix~\ref{app:multi_model} and Appendix~\ref{app:real_3region}, respectively.

\section{Discussion}
SPIRE is a time-resolved shared/private representation model for continuous multi-region LFP. The strongest evidence for its decomposition comes from two complementary tests: complete intended-versus-crossed recovery when ground truth is known, and nonlinear cross-region transfer in held-out human LFP. These tests, together with the distinct spectral organization observed in human
LFP, support a narrower and more defensible interpretation than strict
disentanglement: the shared branch preferentially represents information transferable between recorded regions, whereas the private branch preferentially represents nontransferable regional structure.

Applied to human DBS recordings, this organization reveals a consistent network-level pattern. Shared trajectories shift more strongly under stimulation and encode stimulation condition more accurately than private trajectories. The matched CTAE analysis reaches the same qualitative conclusion, which suggests that the qualitative shared-over-private pattern is not unique
to SPIRE, while SPIRE preserves substantially more stimulation information. These observations are consistent with network perspectives on DBS~\citep{mcintyre2010network,horn2020opportunities} and with theories emphasizing distributed coordination in basal ganglia-thalamo-cortical circuits~\citep{galvan2015alterations,jinnah2006new,obeso2008basal,zhuang2004neuronal}.

Several limitations define the scope of the claims. First, ``shared'' means shared among the \emph{recorded} regions. Unobserved regions or common inputs may influence both GPi and STN, so shared activity cannot be interpreted as direct causal communication. ConvAlign establishes temporal correspondence, not causal direction or information flow. Second, SPIRE learns latent trajectories rather than an explicit dynamical transition law and is not intended for future prediction or simulation of unseen perturbations. Third, the current formulation uses one shared and one private branch per region and does not explicitly separate globally shared from subnetwork-specific factors; with heterogeneous larger recordings, subnetwork structure may be split across the two spaces. Fourth, directed pairwise aligners scale as $O(|\mathcal{R}|^2)$, making the present implementation best suited to a modest number of regions (\ref{app:extension_multi}). Fifth, models are hemisphere specific; learning a common cross-subject latent space remains an unvalidated extension of SPIRE. Functional electrode embeddings that accommodate heterogeneous electrode configurations across subjects~\citep{javadzadeh2025functional} offer a complementary approach toward this broader objective, although integrating such representations with SPIRE would require additional validation. Finally, SPIRE requires dataset-level choices for latent dimensionality,
regularization strengths, and optimization settings. Appendix~\ref{app:external_rat} provides a preliminary application to a
four-animal CA1-PFC dataset after dataset-specific adaptation, while Appendix~\ref{app:robustness} reports no-lag, dimensionality, and alternative-objective robustness analyses.

Overall, the results position SPIRE as a practical shared/private representation tool for continuous multiregion LFP rather than a causal or generative dynamical system. The DBS application illustrates how such a decomposition can separate
perturbation-related changes expressed across recorded regions from changes
that remain region specific.

\section*{AI use statement}

We used OpenAI's ChatGPT as an assistive tool during implementation, analysis,
and manuscript preparation. The research question, model design, objectives,
experimental analyses, and scientific conclusions were developed and determined
by the authors. Generative AI assisted with implementing author-specified model
and analysis code; coding the synthetic-data generator from LFP characteristics
specified by the authors; generating plotting and analysis scripts; suggesting
additional robustness checks; and providing secondary feedback on the
interpretation and presentation of results. It was also used to summarize
selected literature identified by the authors and to assist with manuscript
drafting, editing, organization, and phrasing.

Generative AI was not used for collection or preprocessing of the human neural
recordings. All AI-assisted code and text incorporated into
the work were reviewed and verified by the authors, who made the final
methodological and interpretive decisions and take full responsibility for the
content of the paper.

\section*{Ethics statement}

The human intracranial recordings analyzed in this study were collected under
protocols approved by the institutional review boards of the participating
medical centers, with informed consent obtained in accordance with applicable
regulations. The study analyzes recordings acquired from pediatric patients undergoing
clinically indicated intracranial monitoring, with research stimulation
performed under approved protocols.
Participant information is anonymized in the submission, and the analyses in
this work do not alter clinical treatment decisions. We report results at the
subject/hemisphere or aggregate level and limit our claims to properties of the
recorded signals and learned representations.

\section*{Reproducibility statement}

We provide the information required to reproduce the model and experiments
throughout the paper and supplementary material. The complete SPIRE objective,
architecture, optimization schedules, and multi-region formulation are described
in Appendix~\ref{app:spire_losses}-\ref{app:multi_model}. Synthetic-data
generation and baseline fitting procedures are detailed in
Appendix~\ref{app:datagen}-\ref{app:ctae_fit_synth}. Human-data preprocessing,
latent-dimension selection, and training procedures are provided in
Appendix~\ref{app:real_info}, Appendix~\ref{app:var_sweep}, and
Appendix~\ref{app:human_training}. Additional robustness analyses and complete
recovery results are reported in Appendix~\ref{app:robustness}. The human intracranial recordings contain sensitive clinical data and are not
publicly distributed. De-identified data may be made available upon reasonable
request, subject to completion of an appropriate data-use agreement and any
applicable institutional and ethical approvals. Code and experiment scripts are available in the anonymous repository:
\url{https://github.com/GitAnonymSubmission/spire-ICLR2027}.

\bibliographystyle{iclr2027_conference}
\bibliography{ref}

\renewcommand{\thefigure}{A.\arabic{figure}}
\setcounter{figure}{0}
\renewcommand{\thetable}{A.\arabic{table}}
\setcounter{table}{0}
\renewcommand{\theequation}{A.\arabic{equation}}
\setcounter{equation}{0}

\appendix
\section{Technical appendices and supplementary material}
\subsection{SPIRE details}

\subsubsection{Detailed Training Objectives}
\label{app:spire_losses}
Let $\mathcal{R}$ denote the set of regions. For any region $r\in\mathcal{R}$, let $x^{(r)}\in\mathbb{R}^{B\times T\times C_r}$ be the input
and $\hat{x}^{(r)}$ its reconstruction. Let $z^{(r)}_{\text{sh}}\in\mathbb{R}^{B\times T\times d_{\text{sh}}}$ and
$z^{(r)}_{\text{pr}}\in\mathbb{R}^{B\times T\times d_{\text{pr}}}$ be shared and private latents, and
$\tilde{z}^{(s\to r)}_{\text{sh}} = M^{(s\to r)}\,\mathrm{ConvAlign}(z^{(s)}_{\text{sh}})$ the temporally aligned and linearly mapped
shared latents from $s$ into the space of $r$ (see main text). When needed, we flatten time and batch so that
$Z^{(r)}_{\bullet}\in\mathbb{R}^{(BT)\times d_\bullet}$, write $N{=}BT$, and use
\[
\mathrm{Cov}(A,B) \;=\; \tfrac{1}{N-1}\,A^\top B, \qquad
\mathrm{std}(\cdot)\ \text{computed per feature over the $N$ rows.}
\]
\paragraph{Reconstruction objective.}

The primary reconstruction term ensures that each region can be reconstructed
from its own shared and private latent representations:
\begin{align}
\mathcal{L}_{\mathrm{rec}}
=
\sum_{r\in\mathcal{R}}
\left\|
x^{(r)}-\hat{x}^{(r)}
\right\|_2^2.
\end{align}

To encourage shared latents to encode transferable cross-regional structure,
we additionally include shared-only reconstruction losses.

Cross-reconstruction uses only aligned shared latents from another region:
\begin{align}
\mathcal{L}_{\mathrm{cross}}
=
\sum_{\substack{r,s\in\mathcal{R}\\r\neq s}}
\left\|
x^{(r)}
-
f_{\mathrm{dec}}^{(r)}
(
\tilde{z}^{(s\to r)}_{\mathrm{sh}},0
)
\right\|_2^2.
\end{align}

Self-reconstruction uses only the region’s own shared latents:
\begin{align}
\mathcal{L}_{\mathrm{self}}
=
\sum_{r\in\mathcal{R}}
\left\|
x^{(r)}
-
f_{\mathrm{dec}}^{(r)}
(
z^{(r)}_{\mathrm{sh}},0
)
\right\|_2^2.
\end{align}
\paragraph{Alignment objective.}

Shared latent spaces are aligned across regions using VICReg:
\begin{align}
\mathcal{L}_{\mathrm{align}}
=
\frac12
\sum_{\substack{r,s\in\mathcal{R}\\r\neq s}}
\Big(
\mathrm{VICReg}
(
z^{(r)}_{\mathrm{sh}},
\tilde{z}^{(s\to r)}_{\mathrm{sh}}
)
+
\mathrm{VICReg}
(
z^{(s)}_{\mathrm{sh}},
\tilde{z}^{(r\to s)}_{\mathrm{sh}}
)
\Big).
\end{align}

To preserve interpretability and prevent excessive deformation,
alignment modules are softly regularized toward identity mappings.

Linear mappers are penalized via
\begin{align}
\mathcal{L}_{\mathrm{mapid}}
=
\sum_{s\to r}
\left\|
M^{(s\to r)}-I
\right\|_F^2.
\end{align}

ConvAlign filters are regularized toward impulse-like unit-sum kernels:
\begin{align}
\mathcal{L}_{\mathrm{align-reg}}
=
\sum_{s\to r}
\sum_{j=1}^{d_{\mathrm{sh}}}
\Big(
\|
k_{s\to r,j}-\delta
\|_2^2
+
\|
\mathbf{1}^\top k_{s\to r,j}-1
\|_2^2
\Big).
\end{align}
\paragraph{Disentanglement objective.}

To reduce redundancy between shared and private subspaces,
we penalize the cross-covariance between per-feature standardized latents:
\begin{align}
\mathcal{L}_{\mathrm{orth}}
=
\sum_{r\in\mathcal{R}}
\|
\mathrm{Cov}
(
\bar{Z}^{(r)}_{\mathrm{sh}},
\bar{Z}^{(r)}_{\mathrm{pr}}
)
\|_F^2,
\end{align}
where $\bar{Z}$ denotes zero-mean, unit-variance normalization.

To avoid degenerate solutions, we additionally include variance guards:
\begin{align}
\mathcal{L}_{\mathrm{var-sh}}
&=
\sum_{r\in\mathcal{R}}
\|
\mathrm{std}
(
z^{(r)}_{\mathrm{sh}}
)
-1
\|_2^2,
\\
\mathcal{L}_{\mathrm{var-pr}}
&=
\sum_{r\in\mathcal{R}}
\sum_{j=1}^{d_{\mathrm{pr}}}
\max
\left(
0,
\tau-
\mathrm{std}
(
z^{(r)}_{\mathrm{pr},j}
)
\right)^2.
\end{align}

\paragraph{Total objective.}

The full SPIRE objective is:
\begin{align}
\mathcal{L}
=
\lambda_{\mathrm{rec}}\mathcal{L}_{\mathrm{rec}}^{\mathrm{family}}
+
\lambda_{\mathrm{align}}\mathcal{L}_{\mathrm{align}}^{\mathrm{family}}
+
\lambda_{\mathrm{dis}}\mathcal{L}_{\mathrm{dis}}^{\mathrm{family}},
\end{align}
where

\begin{align}
\mathcal{L}_{\mathrm{rec}}^{\mathrm{family}}
&=
\mathcal{L}_{\mathrm{rec}}
+
\lambda_{\mathrm{self}}\mathcal{L}_{\mathrm{self}}
+
\lambda_{\mathrm{cross}}\mathcal{L}_{\mathrm{cross}},
\\
\mathcal{L}_{\mathrm{align}}^{\mathrm{family}}
&=
\mathcal{L}_{\mathrm{align}}
+
\lambda_{\mathrm{map}}\mathcal{L}_{\mathrm{mapid}}
+
\lambda_{\mathrm{conv}}\mathcal{L}_{\mathrm{align-reg}},
\\
\mathcal{L}_{\mathrm{dis}}^{\mathrm{family}}
&=
\mathcal{L}_{\mathrm{orth}}
+
\lambda_{\mathrm{sh}}\mathcal{L}_{\mathrm{var-sh}}
+
\lambda_{\mathrm{pr}}\mathcal{L}_{\mathrm{var-pr}}.
\end{align}

The reconstruction term was fixed to weight 1 throughout all experiments, and variance-guard coefficients were fixed across datasets. In practice, only a subset of loss coefficients were scheduled during training;
variance guards remained fixed across datasets.

\subsubsection{Implementation Details}
\label{app:spire_impl}

\paragraph{Lag augmentation.}
To improve robustness to temporal misalignment, SPIRE receives lag-augmented features.
For region~1 (GPi) with $C$ channels and sequence length $T$, we stack the signal and its
lagged versions up to lag $L$:
\begin{equation}
\label{eq:lagged_input}
X^{(1)} = 
\begin{bmatrix}
x^{(1)}_{:,\,0:T-L} \\
x^{(1)}_{:,\,1:T-L+1} \\
\vdots \\
x^{(1)}_{:,\,L:T}
\end{bmatrix}
\;\;\in \mathbb{R}^{C(L+1) \times (T-L)} ,
\end{equation}
and analogously for region~2 (STN).
This expands each input into $(L+1)$ lagged copies, enabling SPIRE to model short-range
delays directly.

\paragraph{Training setup.}
Data are split 80/20 into training and validation sets, with mini-batches of size 8 for training and batch size 1 for validation. 
Learning rate is initialized at $10^{-3}$ and halved on plateaus (patience 10 epochs). 
Gradients are scaled and clipped to an L2 norm of 1.0. 
Validation loss is monitored each epoch with early stopping (patience 20, ignoring the first 140 epochs).

\subsubsection{Multi-region extension}
\label{app:multi_model}
\paragraph{Model overview.}

SPIRE generalizes naturally from two-region to multi-region recordings.
Given a set of regions $\mathcal{R}=\{r_1,\dots,r_n\}$, the model instantiates one encoder and one decoder per region. For every ordered pair of distinct regions $(s,r)$, SPIRE additionally instantiates one temporal alignment module $\mathrm{ConvAlign}^{(s\to r)}$ and one linear mapper $M^{(s\to r)}$. Thus, the number of directed alignment pathways scales as $|\mathcal{R}|(|\mathcal{R}|-1)$.

Let
\[
\mathcal{P}=\{(s,r):s,r\in\mathcal{R},\ s\neq r\}
\]
denote the set of ordered region pairs. For each pair, the aligned shared latent is
\[
\tilde{z}_{\mathrm{sh}}^{(s\to r)}
=
M^{(s\to r)}
\mathrm{ConvAlign}^{(s\to r)}
\left(
z_{\mathrm{sh}}^{(s)}
\right).
\]
\paragraph{Multi-region objective.}
The training objective extends directly by averaging region-wise losses over regions and pairwise losses over ordered region pairs. Specifically, reconstruction, self shared-only reconstruction, orthogonality, and variance guards are averaged over $r\in\mathcal{R}$, while cross-reconstruction, shared alignment, mapper identity, and ConvAlign regularization are averaged over $(s,r)\in\mathcal{P}$:
\begin{align}
\mathcal{L}_{\mathrm{rec}}
&=
\frac{1}{|\mathcal{R}|}
\sum_{r\in\mathcal{R}}
\left\|
x^{(r)}-\hat{x}^{(r)}
\right\|_2^2,
\\
\mathcal{L}_{\mathrm{self}}
&=
\frac{1}{|\mathcal{R}|}
\sum_{r\in\mathcal{R}}
\left\|
x^{(r)}
-
f_{\mathrm{dec}}^{(r)}
(
z_{\mathrm{sh}}^{(r)},0
)
\right\|_2^2,
\\
\mathcal{L}_{\mathrm{cross}}
&=
\frac{1}{|\mathcal{P}|}
\sum_{(s,r)\in\mathcal{P}}
\left\|
x^{(r)}
-
f_{\mathrm{dec}}^{(r)}
(
\tilde{z}_{\mathrm{sh}}^{(s\to r)},0
)
\right\|_2^2,
\\
\mathcal{L}_{\mathrm{align}}
&=
\frac{1}{|\mathcal{P}|}
\sum_{(s,r)\in\mathcal{P}}
\mathrm{VICReg}
\left(
z_{\mathrm{sh}}^{(r)},
\tilde{z}_{\mathrm{sh}}^{(s\to r)}
\right),
\\
\mathcal{L}_{\mathrm{orth}}
&=
\frac{1}{|\mathcal{R}|}
\sum_{r\in\mathcal{R}}
\left\|
\mathrm{Cov}
(
\bar{Z}_{\mathrm{sh}}^{(r)},
\bar{Z}_{\mathrm{pr}}^{(r)}
)
\right\|_F^2.
\end{align}

The variance guards are similarly averaged over regions:
\begin{align}
\mathcal{L}_{\mathrm{var-sh}}
&=
\frac{1}{|\mathcal{R}|}
\sum_{r\in\mathcal{R}}
\left\|
\mathrm{std}
(
z_{\mathrm{sh}}^{(r)}
)
-1
\right\|_2^2,
\\
\mathcal{L}_{\mathrm{var-pr}}
&=
\frac{1}{|\mathcal{R}|}
\sum_{r\in\mathcal{R}}
\sum_{j=1}^{d_{\mathrm{pr}}}
\max
\left(
0,
\tau-
\mathrm{std}
(
z_{\mathrm{pr},j}^{(r)}
)
\right)^2.
\end{align}

Finally, alignment-module regularizers are averaged over ordered region pairs:
\begin{align}
\mathcal{L}_{\mathrm{mapid}}
&=
\frac{1}{|\mathcal{P}|}
\sum_{(s,r)\in\mathcal{P}}
\left\|
M^{(s\to r)}-I
\right\|_F^2,
\\
\mathcal{L}_{\mathrm{align-reg}}
&=
\frac{1}{|\mathcal{P}|}
\sum_{(s,r)\in\mathcal{P}}
\mathrm{Reg}
\left(
\mathrm{ConvAlign}^{(s\to r)}
\right).
\end{align}

The total loss is the same weighted combination used in the two-region case:
\begin{align}
\mathcal{L}
=&
w_{\mathrm{rec}}\mathcal{L}_{\mathrm{rec}}
+
w_{\mathrm{align}}\mathcal{L}_{\mathrm{align}}
+
w_{\mathrm{orth}}\mathcal{L}_{\mathrm{orth}}
+
w_{\mathrm{self}}\mathcal{L}_{\mathrm{self}}
+
w_{\mathrm{cross}}\mathcal{L}_{\mathrm{cross}}
\\
&
+
w_{\mathrm{mapid}}\mathcal{L}_{\mathrm{mapid}}
+
w_{\mathrm{align-reg}}\mathcal{L}_{\mathrm{align-reg}}
+
w_{\mathrm{var-sh}}\mathcal{L}_{\mathrm{var-sh}}
+
w_{\mathrm{var-pr}}\mathcal{L}_{\mathrm{var-pr}}.
\end{align}

Because each term is normalized by either $|\mathcal{R}|$ or $|\mathcal{P}|$, adding additional regions does not automatically increase the magnitude of the objective. This normalization allows the same loss schedules to be reused across two-region and three-region SPIRE models and adding a third region does not introduce additional loss-weight hyperparameters.

\subsection{Synthetic data}
\subsubsection{Synthetic benchmark generator}
\label{app:datagen}
Our synthetic generator builds multichannel observations from ground-truth shared and private latent processes. 
\paragraph{Latent dynamics.}
Shared and private latent processes are generated as bursty oscillatory signals with controllable spectral structure. 
Each latent dimension is initialized from a base oscillation with variable frequency, occasional 30-40\,Hz transient bursts, and stochastic frequency shifts over time. 
Optional AR(1) filtering introduces temporal autocorrelation and smooth latent evolution.

To model realistic cross-region nonlinearities, shared latents may undergo region-specific monotone nonlinear warps (\texttt{region\_mismatch}) before observation generation. 
Optional sinusoidal time-varying delays are additionally applied to one region to simulate nonstationary inter-regional timing relationships. 
The generator also supports explicit spectral separation between shared and private processes through independently configurable base frequencies, amplitudes, and colored-noise structure.
\paragraph{Observation generation.}
Each region is assigned a synthetic electrode geometry with contacts arranged in rows. 
Shared and private latent sources are projected to contacts through randomized spatial mixing matrices, optionally incorporating smooth spatial kernels. 
Observed signals are then generated through linear or nonlinear mixing functions (\texttt{gain}, \texttt{bilinear}, or \texttt{gain+bilinear}), with optional multiplicative interactions between shared and private components.

To better approximate intracranial recordings, common-mode structure, row-wise drift, heteroscedastic noise, and sensor noise may be added before bipolar rereferencing is applied.

\paragraph{Noise structure.}
The generator supports multiple biologically inspired noise processes, including colored $1/f$ noise, AR(1) temporal structure, heteroscedastic amplitude-dependent noise, row-wise drift, and low-rank common-mode fluctuations. 
Different benchmark presets selectively enable subsets of these mechanisms to progressively increase realism and modeling difficulty.

\paragraph{Benchmark datasets.}
We evaluated models on four synthetic benchmark regimes of progressively increasing complexity:

\begin{itemize}
    \item \texttt{D0}: approximately linear regime with identity shared mappings, minimal sensor noise, and no temporal misalignment. This setting is favorable to linear latent-variable models such as DLAG.
    
    \item \texttt{D1}: nonlinear regime with region-specific shared warping, nonlinear gain+bilinear mixing, randomized spatial mixing, colored $1/f$ noise, and AR(1) latent dynamics.
    
    \item \texttt{D2}: same as D1, but with an additional sinusoidal time-varying delay between regions to model nonstationary inter-regional timing relationships.
    
    \item \texttt{D3}: spectrally asymmetric regime in which shared latents occupy higher-frequency bands while private latents dominate lower-frequency structure and variance. This setting tests whether models can correctly disentangle cross-regional dynamics from strong region-specific activity.
\end{itemize}

All datasets contain 100 trials of 250 timepoints sampled at 500\,Hz, with 3 shared and 3 private latent dimensions per region.

\begin{figure}[t]
\begin{center}
\includegraphics[width=0.95\linewidth]{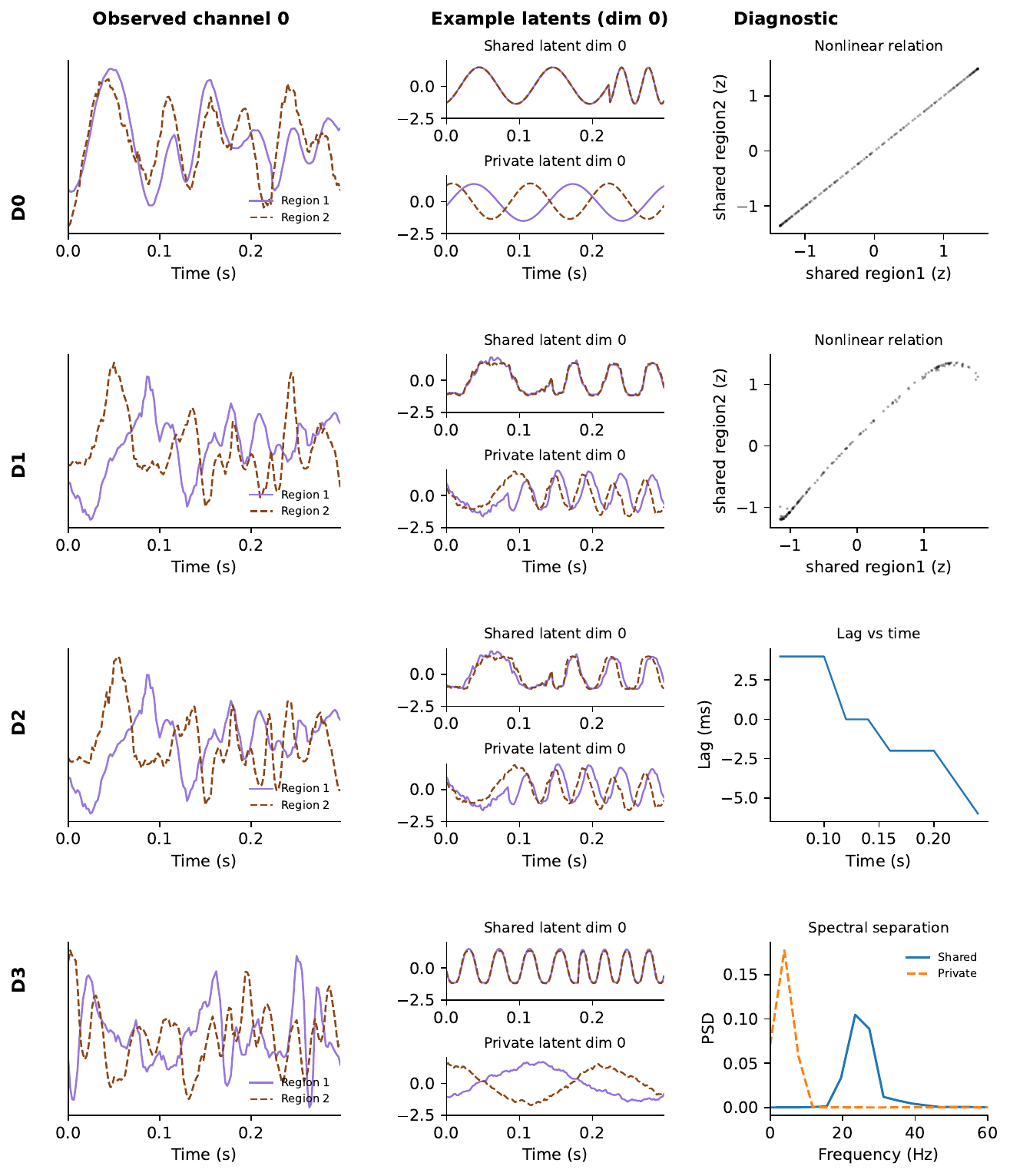}
\end{center}
\caption{
Synthetic benchmark datasets. Each trial is 0.5~s, with a 0.3~s segment shown for visualization. 
Left: example observed channels from region~1 and region~2. 
Middle: example shared and private latent trajectories. 
Right: diagnostic plots showing nonlinear cross-region relationships, time-varying delay structure, and spectral separation between shared and private dynamics.
}
\label{fig:D4synthetic_joint}
\end{figure}

Figure~\ref{fig:D4synthetic_joint} illustrates representative observed channels, latent trajectories, nonlinear cross-region relationships, and time-varying delay structure across the benchmark regimes.

\subsubsection{SPIRE training details}
\label{app:synth_implementation}

\paragraph{Model architecture.}  
Each region used a GRU encoder/decoder (hidden size 64, dropout 0.3). 
Encoders projected into shared ($d_s=3$) and private ($d_p=3$) latents. 
Cross-regional alignment was performed with depthwise ConvAlign filters (kernel size 9) followed by lightweight linear mappers initialized as identity. 
ConvAlign filters were regularized toward impulse responses.  

\paragraph{Training regime.}  
Models were trained for up to 500 epochs. 
A private-gate $\alpha_p$ gradually opened between epochs 80-140, phasing in private latents. 
During epochs 90-110, shared projections were frozen while aligners remained trainable.

\paragraph{Loss scheduling.} 
Loss weights were scheduled in three phases (pre, ramp, post) aligned to the private-gate ramp. 
Table~\ref{tab:loss_schedule} summarizes the schedules for all components.  

\begin{table}[t]
\centering
\caption{Loss weights for synthetic training across phases. 
$\alpha_p$ ramps from 0 to 1 between epochs 80-140.}
\label{tab:loss_schedule}
\begin{tabular}{lccc}
\toprule
\textbf{Loss component} 
& \textbf{Pre (0-80)} 
& \textbf{Ramp (80-140)} 
& \textbf{Post (140-500)} \\
\midrule
$\mathcal{L}_{\text{rec}}$ (reconstruction)        & 1.00  & 1.00  & 1.00 \\
$\mathcal{L}_{\text{align}}$ (shared alignment)     & 0.22  & 0.10  & 0.08 \\
$\mathcal{L}_{\text{cross}}$ (cross recon)          & 0.03  & 0.05  & 0.07 \\
$\mathcal{L}_{\text{self}}$ (self recon)            & 0.02  & 0.04  & 0.03 \\
$\mathcal{L}_{\text{orth}}$ (shared$\perp$private)  & 0.008 & 0.015 & 0.025 \\
$\mathcal{L}_{\text{mapid}}$ (mapper identity)      & 0.010 & 0.005 & 0.000 \\
$\mathcal{L}_{\text{align-reg}}$ (ConvAlign reg.)   & $1\!\times\!10^{-4}$ & $5\!\times\!10^{-4}$ & $1\!\times\!10^{-4}$ \\
\midrule
\multicolumn{4}{l}{\textit{Fixed variance guards}} \\
$\mathcal{L}_{\text{var}}^{\text{shared}}$ (unit var) & 0.005 & 0.005 & 0.005 \\
$\mathcal{L}_{\text{var}}^{\text{private}}$ (floor) & 0.002 & 0.002 & 0.002 \\
\bottomrule
\end{tabular}

\vspace{2pt}
\footnotesize
Freeze window: epochs 90-110 (shared projections frozen, aligners remain trainable).
\end{table}

\subsubsection{DLAG fitting details}
\label{app:dlag_fit}

\paragraph{Data formatting.}
Following the public DLAG demo, we represent each trial as a struct with fields \texttt{trialId}, \texttt{T}, and \texttt{y}. For our two regions (GPi-like = group~1; STN-like = group~2), we construct \texttt{y} by stacking the trial’s multichannel recordings from both regions along the channel dimension, yielding a matrix of shape \([C_1{+}C_2] \times T\) per trial. A small i.i.d.\ Gaussian jitter (\(\varepsilon\sim \mathcal{N}(0, \sigma^2 I)\), \(\sigma\approx 10^{-6}\)) is added to avoid singular covariances during initialization. 
We set the across-group (shared) and within-group (private) latent dimensionalities in DLAG to match the synthetic ground-truth for each preset: 3 shared and 3 private latents.

\paragraph{Temporal and kernel settings.}
We follow the demo configuration and use the RBF Gaussian-process kernel (\(\texttt{covType='rbf'}\)) for latent dynamics. The GP timescale initialization is \(\texttt{startTau}=2\times\texttt{binWidth}\). The \(\texttt{binWidth}\) parameter (ms) is set to the sample period used to bin the synthetic trials, and \(\texttt{segLength}=T\) matches the trial length (in time bins).

\paragraph{Optimization and stopping.}
We fit DLAG with EM up to \(\texttt{maxIters}=500\) (demo-style runs may use higher values), log-likelihood tolerance \(\texttt{tolLL}=10^{-8}\), and progress diagnostics saved every \(\texttt{freqParam}=10\) iterations. Private noise floors are enforced with \(\texttt{minVarFrac}=10^{-3}\). Unless otherwise noted, we do not perform cross-validation (\(\texttt{numFolds}=0\)). We fix a random seed for reproducibility.

\paragraph{Evaluation protocol.}
For each synthetic preset, we train four seeds and align the recovered DLAG latents to ground truth via CCA (performed separately for across-group vs.\ within-group latents). We then report canonical correlations for shared and private components, averaged across regions, trials, and seeds (Appendix Table~\ref{tab:synthetic_latent_recovery}). This evaluation protocol was shared across SPIRE, DLAG, and CTAE to enable consistent comparison of recovered shared and private latent structure across models.

\subsubsection{CTAE training details}
\label{app:ctae_fit_synth}

\paragraph{Data formatting.}
CTAE was trained directly on the synthetic multichannel time-series recordings without modifying the original model architecture. 
For each synthetic preset, region~1 and region~2 recordings were represented as tensors of shape 
\([N \times T \times C]\), where \(N\) denotes trials, \(T\) timepoints, and \(C\) channels. 
Following the original CTAE implementation, the two regions were concatenated along the channel dimension to form a joint multiregion input tensor. 
All channels were z-scored across trials and time prior to training.

To ensure comparability with SPIRE and DLAG, the latent dimensionalities were matched to the synthetic ground truth, using 3 shared and 3 private latent dimensions per region.

\paragraph{Training protocol.}
We used the original CTAE architecture and loss formulation without architectural modifications. 
Training employed the same optimization pipeline used for the real-data experiments, including reconstruction, cross-reconstruction, alignment, and orthogonality objectives. 
Unless otherwise noted, models used 2 transformer layers, 1 attention head, latent dimensionalities matched to ground truth (3 shared and 3 private dimensions), learning rate \(10^{-4}\), and batch size \(8\). 
Positional encodings were enabled using fixed sinusoidal embeddings.

The orthogonality penalty was gradually introduced using a warm-up schedule during early training. 
We additionally explored longer training durations, larger synthetic datasets, and multiple alignment regularization strengths (\(\lambda_{\mathrm{align}}\in\{0.01,0.05\}\)); however, these changes did not substantially alter the qualitative performance trends reported in the main text. 
Early stopping based on validation reconstruction performance was used to prevent overfitting. 

\paragraph{Losses and regularization.}
The CTAE objective combines within-region reconstruction, cross-region reconstruction, shared-latent alignment, and latent orthogonality regularization. 
Orthogonality regularization was gradually increased after an initial warm-up period to stabilize optimization and avoid early collapse of latent representations.

\paragraph{Evaluation protocol.}
For each synthetic benchmark regime, CTAE was trained using four random seeds. 
Recovered shared and private latent trajectories were aligned to the corresponding ground-truth latent processes using canonical correlation analysis (CCA), performed separately for shared and private components. 
Canonical correlations were then averaged across latent dimensions, trials, regions, and seeds. 
This evaluation exactly mirrors the SPIRE protocol to enable direct comparison of disentanglement quality under identical synthetic conditions.

\subsubsection{Ablation analysis}
\label{app:ablation}

To evaluate the contribution of SPIRE’s individual objective families and training-control mechanisms, 
we performed two categories of ablations on the synthetic benchmarks.

\paragraph{Family ablations.}
We first removed entire groups of loss terms:
(i) \texttt{SPIRE\_no\_shared\_only}, which removes shared-only and cross-region reconstruction objectives,
(ii) \texttt{SPIRE\_no\_alignment}, which removes shared-latent alignment and alignment regularization terms, and
(iii) \texttt{SPIRE\_no\_disentangle}, which removes disentanglement-related regularization, including orthogonality and mapper-identity constraints.

\paragraph{Control ablations.}
We additionally evaluated training-control variants:
(i) \texttt{ctrl\_identity\_aligner}, which bypasses convolutional temporal aligners and uses identity alignment,
(ii) \texttt{ctrl\_no\_private\_ramp}, which disables gradual introduction of private latents,
(iii) \texttt{ctrl\_no\_freeze}, which removes the early freeze window during optimization, and
(iv) \texttt{ctrl\_no\_var\_guard}, which disables variance stabilization constraints on latent representations.

All variants were trained using the same protocol as the full SPIRE model across all four synthetic benchmark regimes (D0-D3). 
Recovered shared and private latent trajectories were aligned to ground truth using canonical correlation analysis (CCA), separately for shared and private components. 
Results were averaged across regions, trials, and random seeds. 
Statistical comparisons were performed relative to the full model (\texttt{SPIRE\_synth}) using linear models with Dunnett-adjusted contrasts.

\begin{figure}[t]
  \centering
  \begin{subfigure}[b]{\textwidth}
    \centering
    \includegraphics[width=0.95\linewidth]{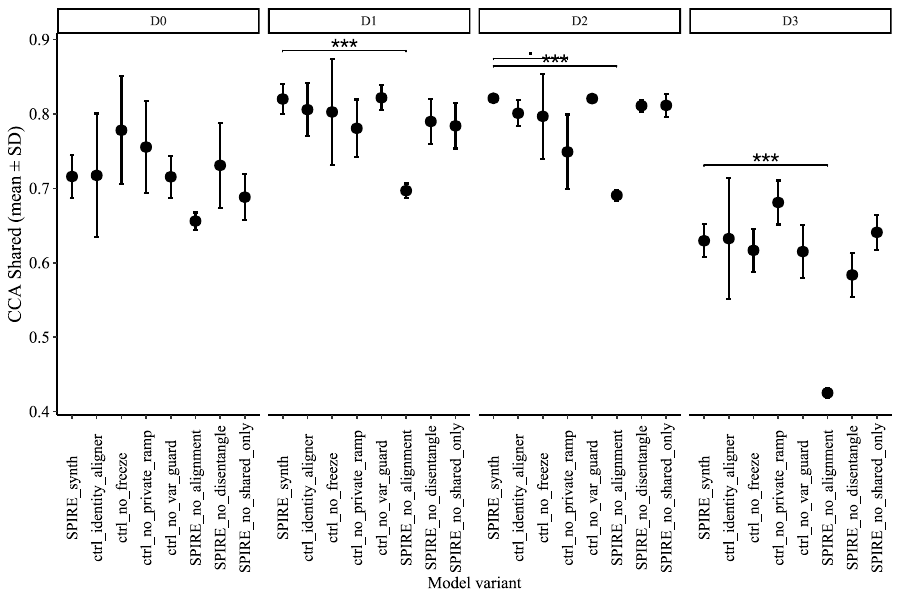}
    \caption{Shared latents}
    \label{fig:ablation_stats_shared}
  \end{subfigure}

  \vspace{1em} 

  \begin{subfigure}[b]{\textwidth}
    \centering
    \includegraphics[width=0.95\linewidth]{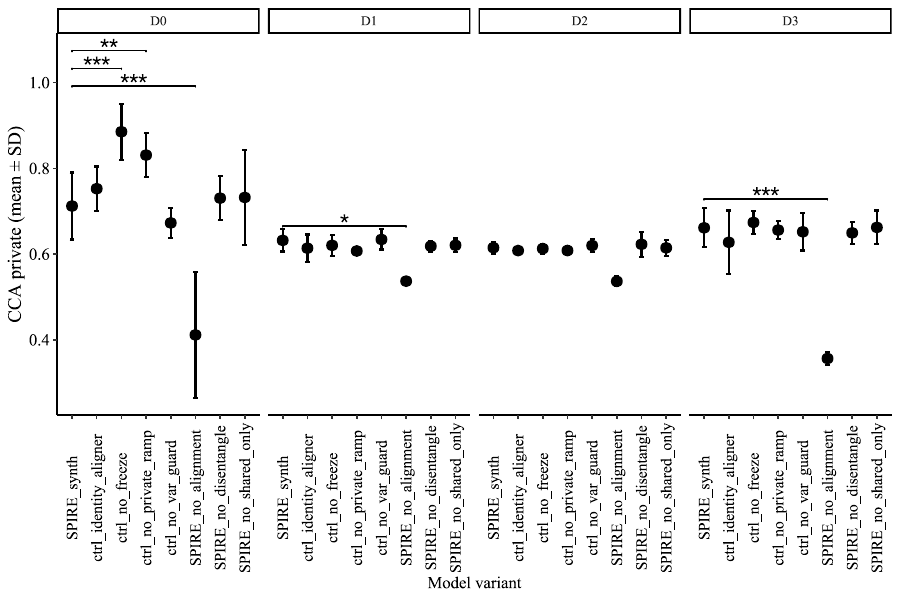}
    \caption{Private latents}
    \label{fig:ablation_stats_private}
  \end{subfigure}

  \caption{Ablation results for shared latents with statistical comparison.
    Mean $\pm$ SD CCA across four seeds, faceted by dataset regime.
    Dunnett-adjusted contrasts are shown relative to the full SPIRE model (\texttt{SPIRE\_synth}).}
  \label{fig:ablation_stats}
\end{figure}

Figures~\ref{fig:ablation_stats_shared} and~\ref{fig:ablation_stats_private} summarize shared and private latent recovery, respectively. 
Across benchmark regimes, no ablation variant consistently outperformed the full SPIRE model. 
Instead, removing any major objective family or control mechanism degraded recovery performance in at least one dataset or latent type.

The impact of ablations varied across synthetic regimes. 
Removing alignment-related terms most strongly degraded shared latent recovery in the nonlinear and temporally distorted settings (D1-D2), highlighting the importance of explicit cross-region alignment under nonlinear warping and time-varying delays. 
Disentanglement-related regularization additionally became critical in the spectrally asymmetric benchmark (D3), where removing orthogonality and mapper constraints substantially reduced both shared and private latent recovery. 

Training-control mechanisms also contributed meaningfully to stable disentanglement. 
Disabling the variance guard produced unstable latent recovery in several regimes, while removing the private latent ramp or optimization freeze window reduced performance selectively depending on dataset complexity. 
Notably, identity alignment occasionally performed competitively in the approximately linear regime (D0), but degraded performance in more realistic nonlinear settings.

Overall, these results demonstrate that SPIRE’s performance depends on the complementary interaction of reconstruction, alignment, disentanglement, and training-control mechanisms. 
Different synthetic regimes stress different aspects of the model, and no reduced variant matched the robustness of the full SPIRE objective across all benchmark conditions.

\subsubsection{Additional CCA recovery and trajectory results}
\label{app:synth_cca_results}

The main text evaluates ground-truth recovery using nonlinear,
cross-validated $R^2$, which can capture nonlinear relationships between
learned representations and the generating latent factors. As a complementary
linear measure, we also quantified recovery using canonical correlation
analysis (CCA). CCA measures similarity between the learned latent subspace and
the corresponding ground-truth latent subspace while being invariant to linear
rotations of the latent coordinates. We additionally visualize representative
trajectories after CCA-based alignment to facilitate qualitative comparison of
the recovered and ground-truth dynamics.

\begin{table}[t]
\centering
\caption{
CCA-based synthetic latent recovery across benchmark regimes.
Values report mean canonical correlation (SD) between learned and corresponding
ground-truth latent subspaces across four matched random seeds. Bold indicates
the highest recovery within each regime and latent class. Significance was
assessed using regime-wise Dunnett-adjusted contrasts comparing each baseline
against \textsc{SPIRE}
($^{\cdot}p<0.1$, $^{*}p<0.05$, $^{**}p<0.01$,
$^{***}p<0.001$).
}
\label{tab:synthetic_latent_recovery}
\begin{tabular}{llccc}
\toprule
Dataset & Metric & SPIRE & DLAG & CTAE \\
\midrule
\multirow{2}{*}{D0}
& Shared latent recovery
& 0.72 (0.03)
& \textbf{0.91 (0.03)}$^{***}$
& 0.26 (0.04)$^{***}$ \\
& Private latent recovery
& 0.71 (0.08)
& \textbf{0.87 (0.05)}$^{***}$
& 0.38 (0.02)$^{***}$ \\
\midrule
\multirow{2}{*}{D1}
& Shared latent recovery
& \textbf{0.82 (0.02)}
& 0.75 (0.01)$^{\cdot}$
& 0.25 (0.09)$^{***}$ \\
& Private latent recovery
& \textbf{0.63 (0.03)}
& 0.55 (0.02)$^{*}$
& 0.27 (0.06)$^{***}$ \\
\midrule
\multirow{2}{*}{D2}
& Shared latent recovery
& \textbf{0.82 (0.00)}
& 0.71 (0.01)$^{**}$
& 0.22 (0.09)$^{***}$ \\
& Private latent recovery
& \textbf{0.61 (0.01)}
& 0.55 (0.01)$^{*}$
& 0.28 (0.05)$^{***}$ \\
\midrule
\multirow{2}{*}{D3}
& Shared latent recovery
& \textbf{0.63 (0.02)}
& 0.47 (0.04)$^{***}$
& 0.21 (0.06)$^{***}$ \\
& Private latent recovery
& \textbf{0.66 (0.05)}
& 0.48 (0.03)$^{***}$
& 0.44 (0.03)$^{***}$ \\
\bottomrule
\end{tabular}
\end{table}

The CCA analysis is consistent with the nonlinear recovery results in the main
text. DLAG achieves the strongest recovery in the approximately linear D0
positive-control regime, whereas \textsc{SPIRE} achieves the strongest shared
and private recovery in D1-D3 as nonlinear observation structure, temporal
misalignment, and spectral confounding are introduced.

\begin{figure*}[t]
  \centering
  \begin{subfigure}[b]{0.48\textwidth}
    \centering
    \includegraphics[width=\textwidth]{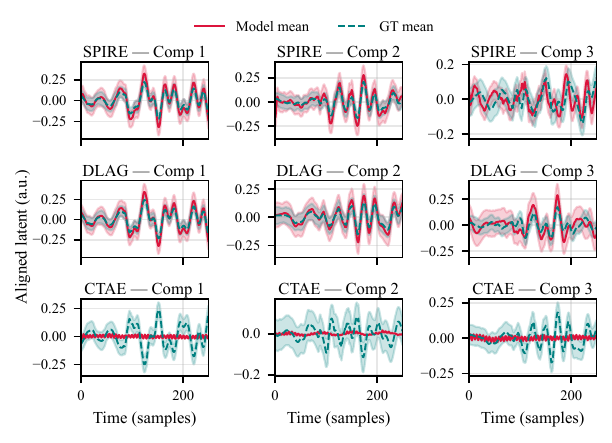}
    \caption{Shared latent recovery.}
  \end{subfigure}
  \hfill
  \begin{subfigure}[b]{0.48\textwidth}
    \centering
    \includegraphics[width=\textwidth]{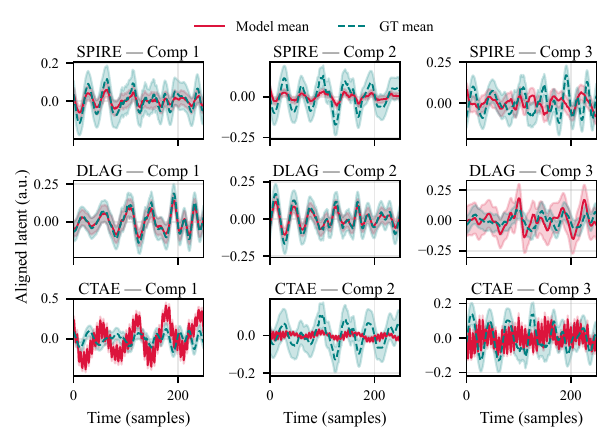}
    \caption{Private latent recovery.}
  \end{subfigure}
  \caption{
Representative latent recovery in the D0 synthetic regime.
Recovered trajectories (solid red) are compared with ground-truth trajectories
(dashed teal) after CCA-based alignment for visualization. Shaded regions
denote SEM across trials.
}
  \label{fig:synth_alignment_example_D0}
\end{figure*}

\begin{figure*}[t]
  \centering
  \begin{subfigure}[b]{0.48\textwidth}
    \centering
    \includegraphics[width=\textwidth]{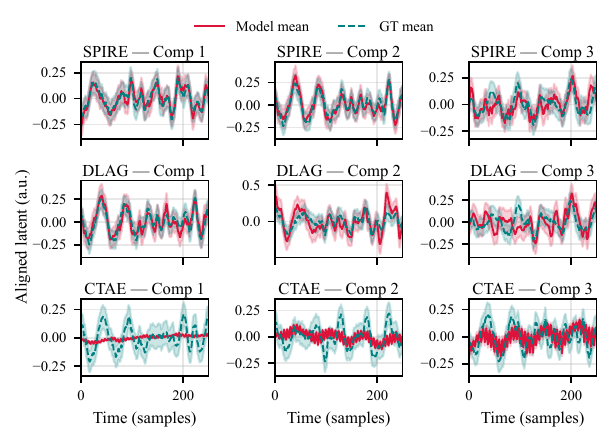}
    \caption{Shared latent recovery.}
  \end{subfigure}
  \hfill
  \begin{subfigure}[b]{0.48\textwidth}
    \centering
    \includegraphics[width=\textwidth]{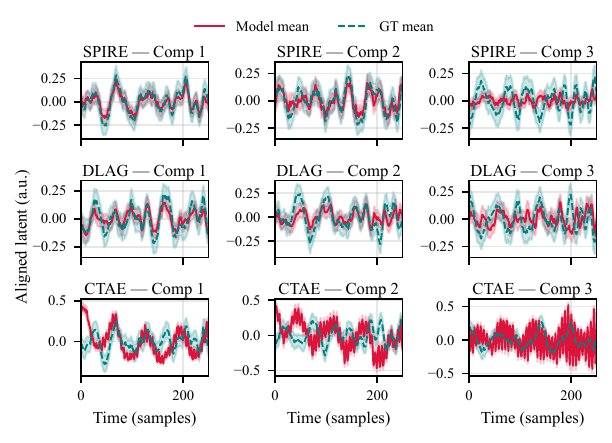}
    \caption{Private latent recovery.}
  \end{subfigure}
  \caption{
Representative latent recovery in the D1 synthetic regime.
Recovered trajectories (solid red) are compared with ground-truth trajectories
(dashed teal) after CCA-based alignment for visualization. Shaded regions
denote SEM across trials.
}
  \label{fig:synth_alignment_example_D1}
\end{figure*}

\begin{figure*}[t]
  \centering
  \begin{subfigure}[b]{0.48\textwidth}
    \centering
    \includegraphics[width=\textwidth]{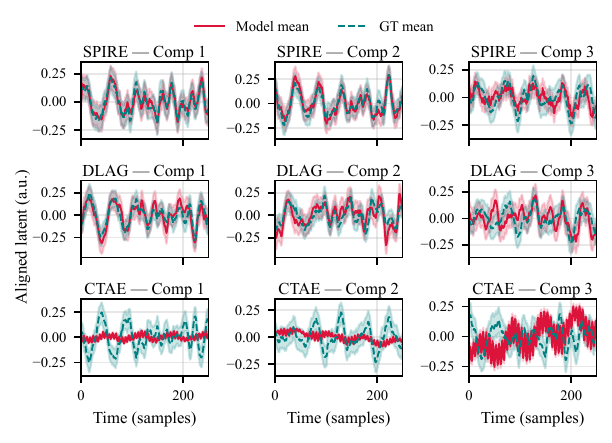}
    \caption{Shared latent recovery.}
  \end{subfigure}
  \hfill
  \begin{subfigure}[b]{0.48\textwidth}
    \centering
    \includegraphics[width=\textwidth]{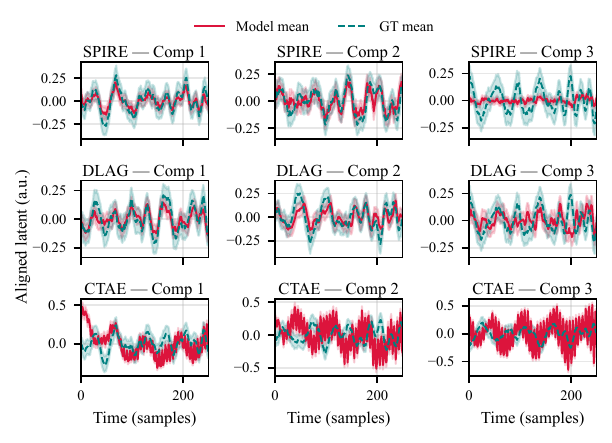}
    \caption{Private latent recovery.}
  \end{subfigure}
  \caption{
Representative latent recovery in the D2 synthetic regime.
Recovered trajectories (solid red) are compared with ground-truth trajectories
(dashed teal) after CCA-based alignment for visualization. Shaded regions
denote SEM across trials.
}
  \label{fig:synth_alignment_example_D2}
\end{figure*}

Representative trajectory recovery for the remaining D0-D2 regimes is shown
in Figs.~\ref{fig:synth_alignment_example_D0}-\ref{fig:synth_alignment_example_D2}.

\subsection{Real intracranial DBS recordings}
\subsubsection{Dataset information}
\label{app:real_info}
\paragraph{Participant information}
Table~\ref{tab:contacts_per_subject} summarizes demographics, etiologies, and the number of usable recording and stimulation contacts per hemisphere. Etiologies spanned genetic, metabolic, and acquired forms of dystonia. All subjects were male and between 5 and 23 years of age.  

Analyses were performed at the hemisphere level. Based on channel quality thresholds (at least three usable microelectrode contacts per region), 17 hemispheres were included for GPi stimulation experiments and 13 for STN stimulation. Due to clinical time constraints, the 250~Hz condition was not administered in 8 hemispheres with STN stimulation hence we only consider 85 and 185 Hz stimulation for STN stimulation analysis.  

\begin{table}
  \caption{Participant demographics and number of stimulation and recording contacts per hemisphere. The table includes each subject's etiology, the number of usable microelectrode recording contacts per region (GPi, STN), and whether GPi or STN stimulation was applied in each hemisphere. Abbreviations: MEPAN = Mitochondrial Enoyl CoA Reductase Protein-Associated Neurodegeneration~\cite{heimer2019mecr}, KMT2B = Lysine Methyltransferase-2B~\cite{Abela1993}, GA1 = Glutaric aciduria type 1~\cite{Boy2017}, PKAN = Pantothenate Kinase-Associated Neurodegeneration~\cite{gregory2017pantothenate}, HIE = Hypoxic-ischemic encephalopathy~\cite{vannucci2000hypoxic}, MYH2 = Myosin heavy chain IIa mutation~\cite{tajsharghi2005mutations}, CP = Cerebral palsy~\cite{krigger2006cerebral}, BG = Basal ganglia.}
  \label{tab:contacts_per_subject}
  \centering
  \resizebox{\textwidth}{!}{%
  \begin{tabular}{l l l c c c c c c c}
    \toprule
    Subject & Etiology & Hemisphere & GPi rec. & STN rec. & VO rec. & VA rec. & VIM rec. & GPi stim. & STN stim. \\
    \midrule
    S1  & MEPAN              & Left  & 18  & 3  & 4 & 4 & 0 & 3 & 1 \\
        &                    & Right & 12  & 6  & 2 & 3 & 0 & 2 & 1 \\
        \midrule
    S2  & GA1                & Left  & 6  & 6  & 1 & 6 & 0 & 2 & 1 \\
        &                    & Right & 9  & 1  & 1 & 4 & 0 & 2 & 1 \\
        \midrule
    S3  & Atypical PKAN      & Left  & 11 & 4  & 1 & 0 & 0 & 4 & 1 \\
        &                    & Right & 8 & 4  & 1 & 0 & 0 & 3 & 1 \\
        \midrule
    S4  & HIE                & Left  & 6  & 3  & 1 & 5 & 0 & 2 & 0 \\
        &                    & Right & 3  & 3  & 2 & 7 & 0 & 2 & 0 \\
        \midrule
    S5  & CP (Prematurity)   & Left  & 3  & 1  & 1 & 0 & 0 & 1 & 1 \\
        &                    & Right & 3  & 4  & 1 & 0 & 0 & 1 & 2 \\
        \midrule
    S6  & GA1                & Left  & 3  & 4  & 1 & 2 & 0 & 2 & 1 \\
        &                    & Right & 6  & 0  & 2 & 2 & 0 & 2 & 1 \\
        \midrule
    S7 & Kernicterus~\cite{hamza2019kernicterus} & Left  & 9  & 3  & 0 & 0 & 3 & 2 & 1  \\
       &                                           & Right & 8  & 4  & 1 & 0 & 3 & 2 &  2\\
       \midrule
    S8 & BG hemorrhage      & Left  & 6  & 3  & 4 & 2 & 0 & 1 & 0 \\
       &                    & Right & 9  & 3  & 3 & 2 & 0 & 1 & 0 \\
       \midrule
    S9 & KMT2B              & Left  & 7 & 4 & 1 & 0 & 4 & 3 & 2 \\
       &                    & Right & 7 & 4 & 1 & 0 & 3 & 3 & 2 \\
       \midrule
    S10 & HIE               & Left  & 8  & 4  & 1 & 0 & 4 & 2 & 2 \\
        &                   & Right & 8  & 6  & 2 & 0 & 4 & 2 & 1 \\
    \bottomrule
  \end{tabular}
  }
\end{table}

\paragraph{Stimulation paradigm}
All procedures were approved by institutional review boards at participating medical centers (Children's Hospital of Orange County (CHOC) and Children's Hospital Los Angeles (CHLA)), and informed consent was obtained in accordance with HIPAA regulations.

Stimulation experiments were conducted 2–3 days post-implantation, while temporary stereoelectroencephalography (sEEG) leads remained in place. Our stimulation protocol (adapted from prior work~\citep{hernandez2023effect}) was designed to probe the frequency-dependent effects of DBS using clinical stimulation ranges commonly used for dystonia~\citep{soroushmojdehi2025frequency}. Trains of biphasic pulses were delivered at 85, 185, and 250~Hz through adjacent macro-contacts on each DBS lead. Each frequency–contact pair was tested under two configurations (cathode–anode and reversed polarity), with approximately 1000 pulses delivered per condition (pulse width: 90~µs).  

Amplitude was set to 3~V by default but reduced if subjects experienced discomfort. For frequencies above 100~Hz, a ramping protocol gradually increased voltage from 0 to 3~V over 12 seconds to improve tolerance. Intracranial signals from all implanted leads were recorded simultaneously, with recording duration adjusted to ensure that 1000 pulses were captured in each case.

\paragraph{Signal pre-processing (extended details)}
Recordings were acquired at 24,414~Hz using a Tucker-Davis Technologies (TDT) system and preprocessed initially in MATLAB (R2023b). Bipolar derivation (subtracting adjacent contacts) was applied to enhance spatial specificity and reduce common-mode noise~\citep{mercier2022advances}. To reduce dimensionality while retaining relevant neural dynamics, data were downsampled to 500~Hz, notch filtered and band-limited with a 50~Hz low-pass Butterworth filter. This preprocessing removed high-frequency stimulation artifacts~\citep{alarie2022artifact} while preserving oscillatory components of interest.

Then signals were segmented into non-overlapping 0.5-s windows. To capture temporal dependencies, lag-augmented features with 0–3 sample delays were appended to each channel, expanding the feature dimensionality four-fold. This choice balanced computational efficiency with the ability to represent short-timescale spatiotemporal dynamics.  

Finally, data were split into training (80\%) and test (20\%) sets for each subject and hemisphere before model fitting.

\subsubsection{Variance partition and model selection}
\label{app:var_sweep}

\paragraph{Motivation.}
Subjects varied widely in etiology, anatomy, and recording coverage, so we allowed latent dimensionalities to adapt per subject–hemisphere. We trained SPIRE across a grid of shared (3–5) and private (2–4) latent dimensions. To compare candidate models, we quantified how much observed variance in GPi and STN signals was uniquely explained by shared latents, uniquely explained by private latents, or redundantly captured by both. This partitioning ensured that both subspaces contributed meaningfully, avoiding degenerate solutions where all variance collapses into one.

\paragraph{Fraction of variance explained (FVE).}
For each region $r \in \{\text{GPi}, \text{STN}\}$, we computed cross-validated fraction of variance explained (FVE) by regressing observed signals $Y^r$ on latent trajectories $Z$:
\[
\mathrm{FVE}(Z \rightarrow Y^r) = 1 - \frac{\sum \| Y^r - \hat{Y}^r(Z) \|^2}{\sum \| Y^r - \bar{Y}^r \|^2},
\]
where $\hat{Y}^r(Z)$ is the ridge-regression prediction and $\bar{Y}^r$ the global mean.  
We evaluated three cases:
\[
\mathrm{FVE}_S^r = \mathrm{FVE}(Z_S^r \!\rightarrow\! Y^r), \quad
\mathrm{FVE}_P^r = \mathrm{FVE}(Z_P^r \!\rightarrow\! Y^r), \quad
\mathrm{FVE}_{SP}^r = \mathrm{FVE}([Z_S^r,Z_P^r] \!\rightarrow\! Y^r).
\]

\paragraph{Order-free partition.}
Using these, we derived an order-invariant variance partition:
\begin{align*}
\mathrm{UniqueShared}^r &= \max\!\big(0, \; \mathrm{FVE}_{SP}^r - \mathrm{FVE}_P^r \big),\\
\mathrm{UniquePrivate}^r &= \max\!\big(0, \; \mathrm{FVE}_{SP}^r - \mathrm{FVE}_S^r \big).
\end{align*}
\[
\mathrm{Redundant}^r = \max\!\big(0, \; \mathrm{FVE}_S^r + \mathrm{FVE}_P^r - \mathrm{FVE}_{SP}^r \big).
\]

\paragraph{Model selection rule.}
\label{app:selection_rule}
We trained SPIRE across $(d_s,d_p)\in\{3,4,5\}\times\{2,3,4\}$ per subject–hemisphere. Variants were filtered by:  
(i) redundancy threshold $\max_r \mathrm{Redundant}^r \le \tau_{\mathrm{red}}=0.20$,  
(ii) collapse guard $\min_r(\mathrm{UniqueShared}^r,\mathrm{UniquePrivate}^r) \ge \tau_{\mathrm{uni}}=0.01$.  
Among survivors, the variant with lowest validation reconstruction loss was chosen, preferring smaller $(d_s,d_p)$ in ties. 

\begin{figure}[t]
  \centering
  \includegraphics[width=\textwidth]{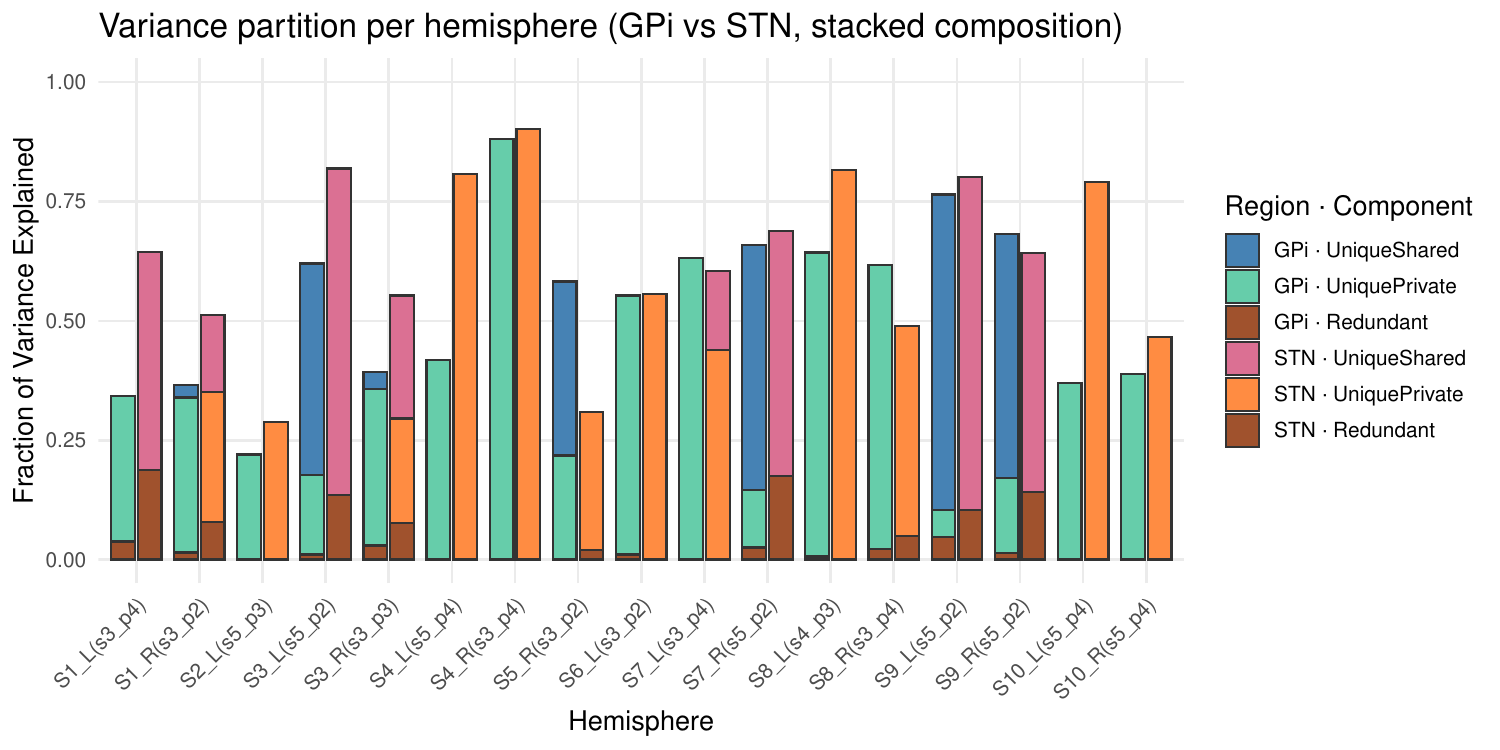}
  \caption{Subject-specific variance partition of GPi vs.\ STN.
  For each subject-hemisphere, two adjacent bars (GPi, STN) show the fraction of variance explained by the \emph{unique shared}, \emph{unique private}, and \emph{redundant} components. Labels include the selected latent sizes $(d_s,d_p)$ per hemisphere chosen via the appendix rule~\ref{app:selection_rule}). Cohort heterogeneity is evident: several hemispheres are dominated by private variance (weak GPi–STN shared structure), whereas others show a more balanced decomposition consistent with stronger cross-regional coordination. Full dimension sweeps (per hemisphere and region) are in Appendix Fig.~\ref{fig:var_sweep}.} 
  \label{fig:var_stack}
\end{figure}

\begin{figure*}[t]
  \centering
  \begin{subfigure}[b]{0.52\textwidth}
    \centering
    \includegraphics[width=\textwidth]{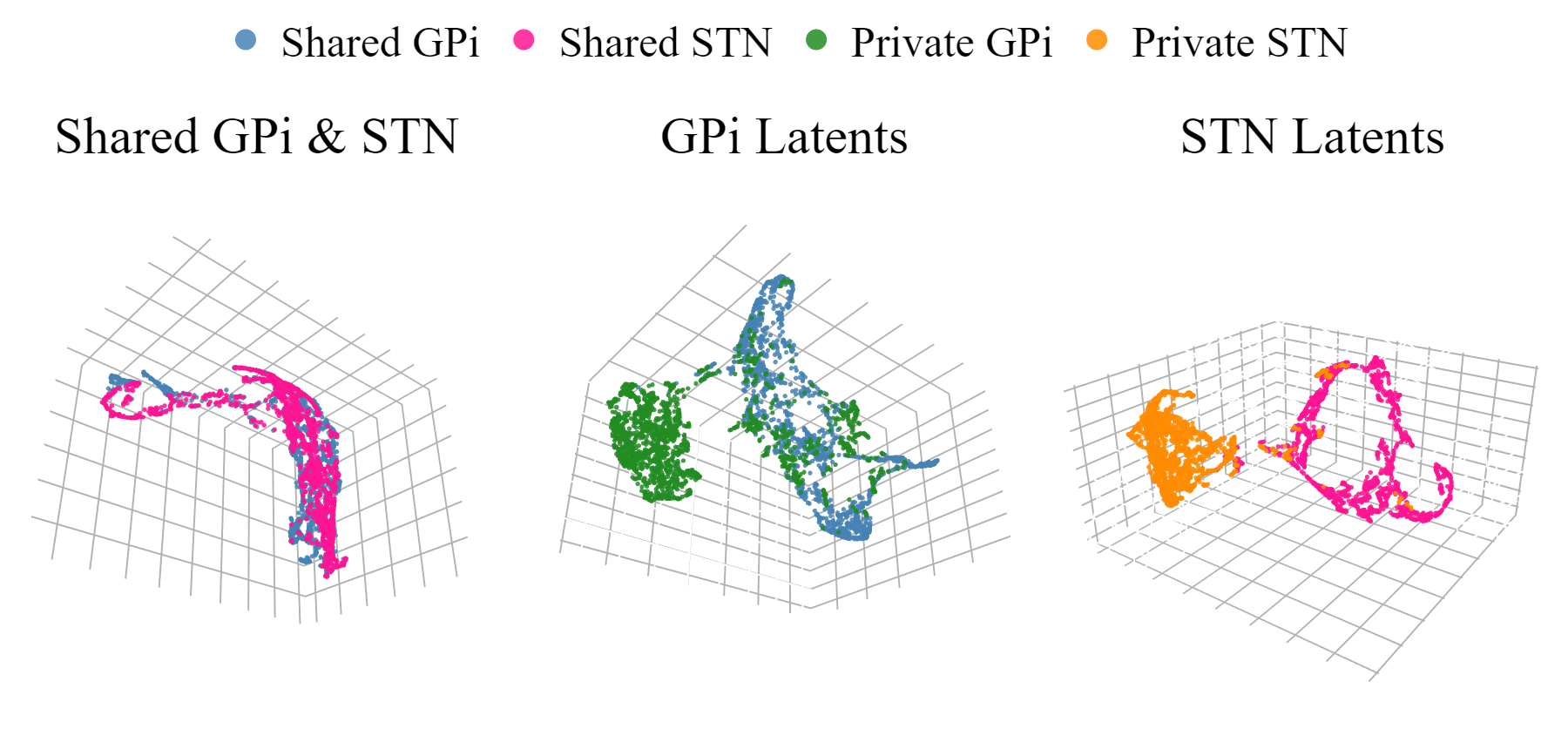}
    \caption{UMAP of shared vs.\ private latents.}
  \end{subfigure}
  \hfill
  \begin{subfigure}[b]{0.46\textwidth}
    \centering
    \includegraphics[width=\textwidth,clip=true,trim= 0in 0in 0in 0in]{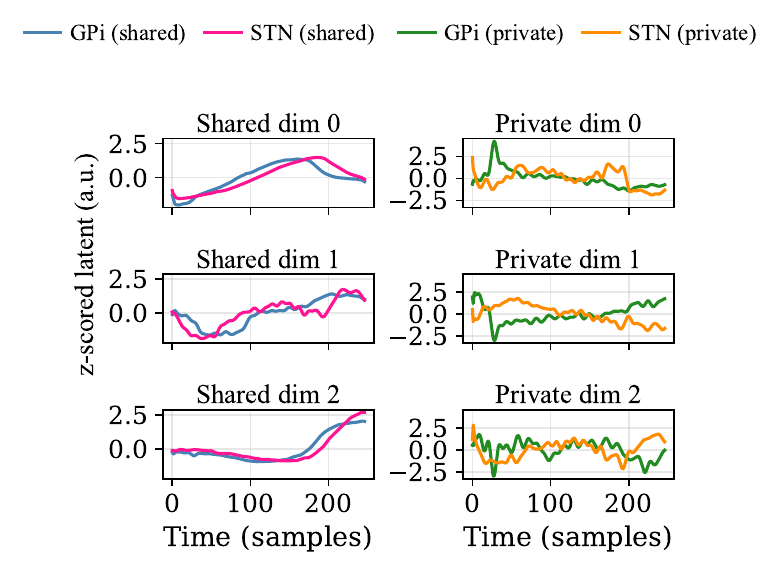}
    \caption{Time-domain latents (shared low-amplitude with lag; private higher variance).}
  \end{subfigure}
\caption{
Representative shared and private latent organization.
(a) 3D UMAPs of shared and private GPi/STN latent spaces.
(b) Corresponding latent trajectories across time.
Shared GPi/STN trajectories are \emph{phase-aligned} but have \emph{very low amplitude} and primarily reflect \emph{slow baseline co-modulation}, whereas private trajectories carry larger, region-specific dynamics that dominate the variance.
}
  \label{fig:subject_examples_umap_time_privatedom}
\end{figure*}

\begin{figure}[t]
  \centering

  \begin{subfigure}[b]{\textwidth}
    \centering
    \includegraphics[width=0.8\textwidth,clip=true,trim= 0in 0in 0in 0.3in]{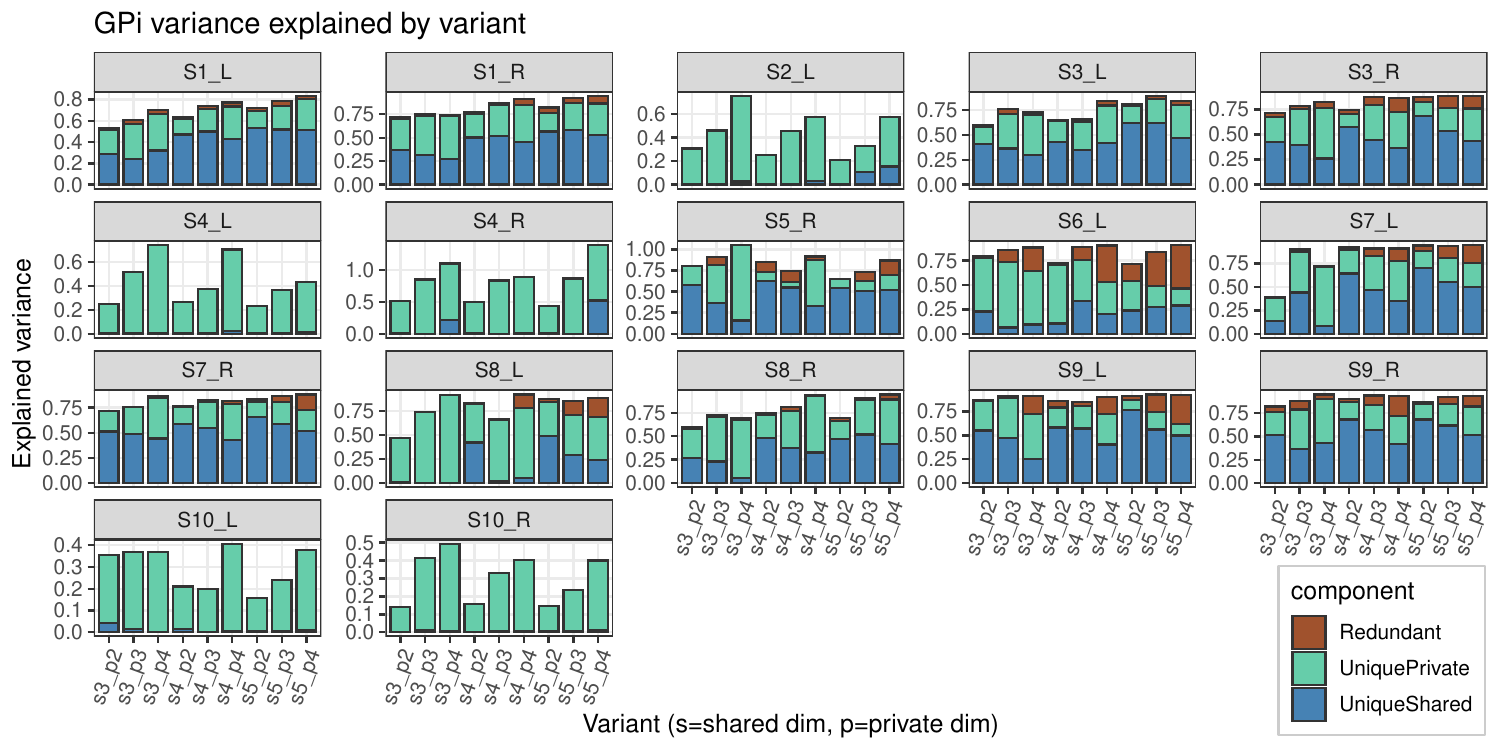}
    \caption{GPi}
    \label{fig:var_sweep_gpi}
  \end{subfigure}

  \vspace{0.8em} 

  \begin{subfigure}[b]{\textwidth}
    \centering
    \includegraphics[width=0.8\textwidth,clip=true,trim= 0in 0in 0in 0.3in]{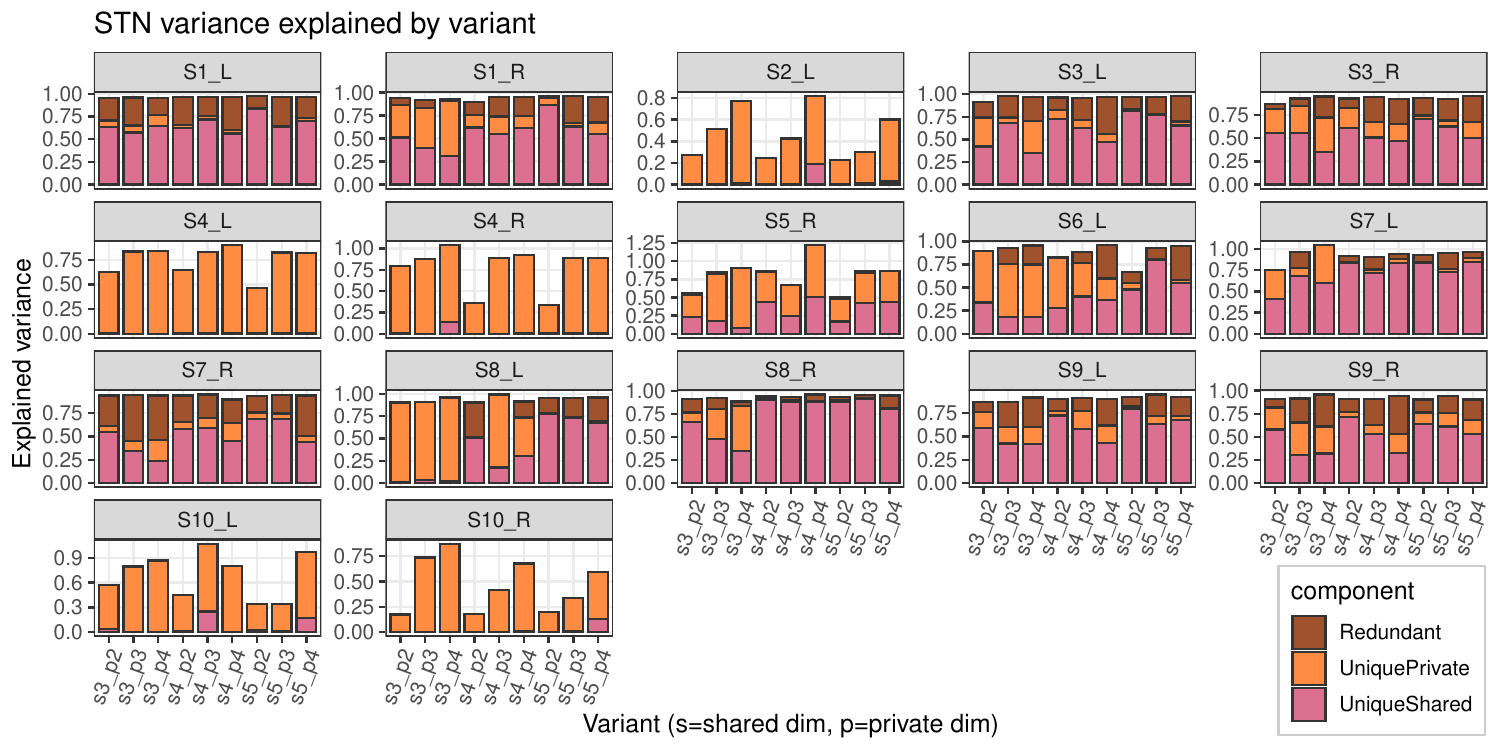}
    \caption{STN}
    \label{fig:var_sweep_stn}
  \end{subfigure}

  \caption{Variance partition across latent dimensionalities.
  For each subject–hemisphere, we trained SPIRE with shared dimension $d_s \in \{3,4,5\}$ and private dimension $d_p \in \{2,3,4\}$. Each bar shows the fraction of explained variance in the observed signal that is uniquely attributable to shared latents, uniquely attributable to private latents, or redundantly captured by both.  
  (a) GPi recordings. (b) STN recordings.  
  These sweeps were used to select the final $(d_s,d_p)$ configuration per subject–hemisphere, as reported in the main text (Fig.~\ref{fig:var_stack}).}
  \label{fig:var_sweep}
\end{figure}

\paragraph{Results.}
Across subjects with chosen dimensions, GPi latents explained $0.302(\pm0.270)$ unique shared and $0.387(\pm0.215)$ unique private variance; STN latents explained $0.401(\pm0.273)$ and $0.403(\pm0.293)$, respectively. Figure~\ref{fig:var_stack} summarizes the chosen models: each subject–hemisphere shows GPi and STN variance partitioned into UniqueShared, UniquePrivate, and Redundant components. The cohort was heterogeneous: some hemispheres (e.g., S8\_R) were private-dominated, while others (e.g., S3) exhibited balanced shared and private contributions. Example UMAP~\citep{mcinnes2018umap} and time-domain traces for balanced case is shown in Appendix Fig.~\ref{fig:subject_examples_umap_time_app} and an example of private dominant case is illustrated in Fig.~\ref{fig:subject_examples_umap_time_privatedom}. Full dimensionality sweeps appear in Fig.~\ref{fig:var_sweep}.

\begin{figure}[h]
  \centering
  \begin{subfigure}[b]{0.52\linewidth}
    \centering
    \includegraphics[width=\textwidth,clip=true,trim=0in 0in 2in 0in]{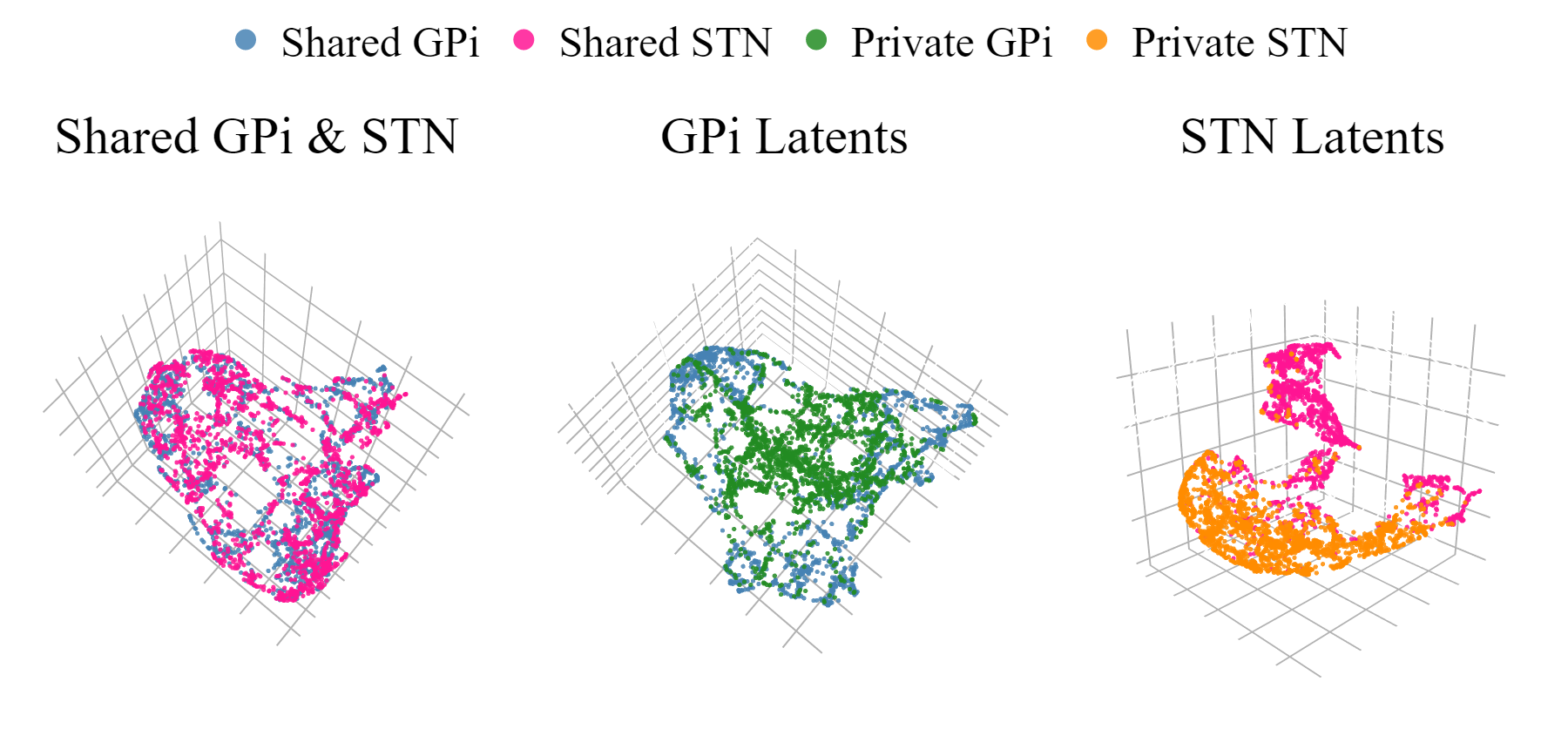}
    \caption{UMAP of shared vs. private latents.}
  \end{subfigure}
  \hfill
  \begin{subfigure}[b]{0.46\linewidth}
    \centering
    \includegraphics[width=\textwidth]{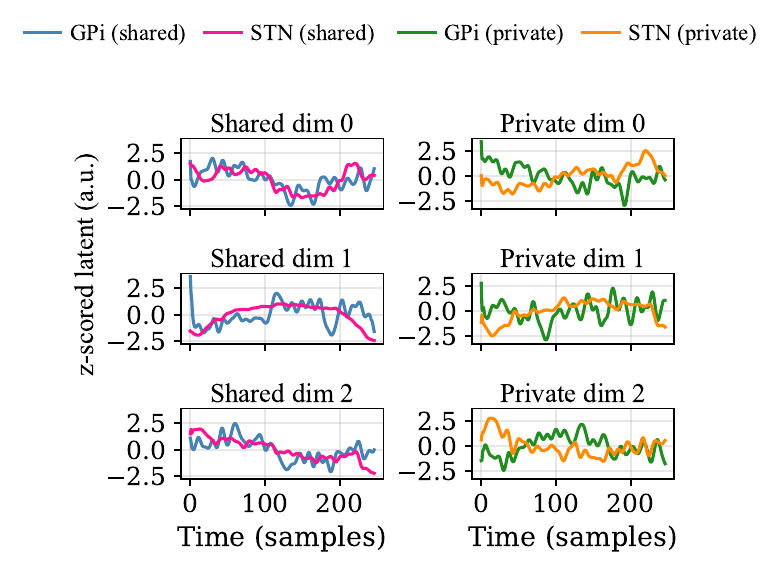}
    \caption{Time-domain latents.}
  \end{subfigure}
  \caption{Representative held-out shared/private latent organization. Shared GPi and STN trajectories overlap more strongly than their private counterparts.}
  \label{fig:subject_examples_umap_time_app}
\end{figure}

\subsubsection{Training Details}
\label{app:human_training}
\paragraph{Architecture.} 
Same GRU encoder–decoder backbone as in the synthetic case, but with dropout 0.2.  
Aligners were initialized as identity mappers with convolutional filters set to impulses.  
\paragraph{Training regime.}  
Models were trained for up to 200 epochs.  
For the first 60 epochs, aligners remained in identity mode (no convolutional shifts); convolutional alignment was enabled thereafter.  
Unlike the synthetic setting, no explicit freeze window was used beyond this identity–to–aligner transition.   
\paragraph{Loss scheduling.}  
Real-data training followed a four-stage schedule tuned for stability of intracranial recordings.  
- Epochs $<60$: warm-up with mapper-only alignment, no cross-reconstruction.  
- Epochs 60–100: enable convolutional alignment, modest self/cross reconstruction.  
- Epochs 100–140: increase alignment and cross/self terms.  
- Epochs $\geq 140$: final tightening of alignment, mapper penalty removed.  Loss weights are reported in Table~\ref{tab:real_loss_schedule}
\begin{table}[t]
\centering
\caption{Loss weight schedule for real DBS training. Values are weight multipliers in the objective.}
\label{tab:real_loss_schedule}
\begin{tabular}{lcccc}
\toprule
\textbf{Loss component} & \textbf{Epoch ($<60$)} & \textbf{(60-100)} & \textbf{(100-140)} & \textbf{($\geq140$)} \\
\midrule
Reconstruction ($\mathcal{L}_{\text{rec}}$) & 1.00 & 1.00 & 1.00 & 1.00 \\
Alignment ($\mathcal{L}_{\text{align}}$)   & 0.30 & 0.30 & 0.38 & 0.45 \\
Orthogonality ($\mathcal{L}_{\text{orth}}$) & 0.012 & 0.012 & 0.015 & 0.015 \\
Self recon. ($\mathcal{L}_{\text{self}}$)  & 0.03 & 0.05 & 0.05 & 0.05 \\
Cross recon. ($\mathcal{L}_{\text{cross}}$) & 0.00 & 0.05 & 0.06 & 0.06 \\
Mapper identity ($\mathcal{L}_{\text{mapid}}$) & 0.005 & 0.005 & 0.003 & 0.000 \\
Aligner regularizer ($\mathcal{L}_{\text{align-reg}}$) & 0.000 & $5\times10^{-5}$ & $7.5\times10^{-5}$ & $1\times10^{-4}$ \\
\midrule
\multicolumn{4}{l}{\textit{Fixed variance guards}} \\
Shared variance guard ($\mathcal{L}_{\text{var}}^{\text{sh}}$) & 0.005 & 0.005 & 0.005 & 0.005 \\
Private variance floor ($\mathcal{L}_{\text{var}}^{\text{pr}}$) & 0.002 & 0.002 & 0.002 & 0.002 \\
\bottomrule
\end{tabular}
\end{table}

\subsubsection{Three-region SPIRE implementation}
\label{app:real_3region}

For the three-region experiments, SPIRE was instantiated with GPi, STN, and thalamic inputs. 
The model used one GRU encoder-decoder pair per region and one directed alignment pathway for each ordered region pair:
GPi$\to$STN, STN$\to$GPi, GPi$\to$Thal, Thal$\to$GPi, STN$\to$Thal, and Thal$\to$STN.
Each directed pathway contained a ConvAlign module followed by an identity-initialized linear mapper.
As in the two-region setting, all region-wise losses were averaged over regions and all pairwise losses were averaged over ordered pairs, preventing the total objective scale from increasing with the number of regions.

\paragraph{Architecture.}
The three-region model used the same GRU encoder-decoder backbone as the two-region real-data model, with hidden size 64, dropout 0.2, ConvAlign kernel size 9, and two decoder layers. 
Shared dimensionality was selected from $d_{\mathrm{sh}}\in\{3,4,5\}$, and private dimensionality was selected from $d_{\mathrm{pr}}\in\{2,3,4\}$ using the same validation-loss and variance-partition criteria used for the two-region models.

\paragraph{Training regime.}
Models were trained for up to 200 epochs using Adam with learning rate $10^{-3}$, weight decay $10^{-5}$, batch size 8, validation batch size 1, mixed precision, gradient clipping at norm 1.0, and ReduceLROnPlateau scheduling. 
Early stopping used patience 20, ignoring the first 140 epochs. 
Unlike the synthetic setting, no explicit shared-latent freeze window was used. 
For the first 60 epochs, alignment was restricted to identity mapping; after epoch 60, ConvAlign modules were enabled.

\paragraph{Loss scheduling.}
Three-region real-data training followed a four-stage schedule. 
Epochs $<60$ used identity alignment and no cross-reconstruction. 
Epochs 60-110 enabled ConvAlign with modest cross-reconstruction. 
Epochs 110-150 increased alignment and cross/self reconstruction. 
Epochs $\geq150$ further tightened alignment while removing the mapper identity penalty. 
Loss weights are reported in Table~\ref{tab:real_3region_loss_schedule}.

\begin{table}[t]
\centering
\caption{Loss weight schedule for three-region SPIRE training on real DBS recordings. Values are weight multipliers in the objective.}
\label{tab:real_3region_loss_schedule}
\begin{tabular}{lcccc}
\toprule
\textbf{Loss component} 
& \textbf{Epoch ($<60$)} 
& \textbf{(60-110)} 
& \textbf{(110-150)} 
& \textbf{($\geq150$)} \\
\midrule
Reconstruction ($\mathcal{L}_{\text{rec}}$) 
& 1.00 & 1.00 & 1.00 & 1.00 \\

Alignment ($\mathcal{L}_{\text{align}}$)   
& 0.22 & 0.24 & 0.30 & 0.36 \\

Orthogonality ($\mathcal{L}_{\text{orth}}$) 
& 0.012 & 0.012 & 0.015 & 0.015 \\

Self recon. ($\mathcal{L}_{\text{self}}$)  
& 0.03 & 0.04 & 0.05 & 0.05 \\

Cross recon. ($\mathcal{L}_{\text{cross}}$) 
& 0.00 & 0.025 & 0.04 & 0.05 \\

Mapper identity ($\mathcal{L}_{\text{mapid}}$) 
& 0.005 & 0.005 & 0.003 & 0.000 \\

Aligner regularizer ($\mathcal{L}_{\text{align-reg}}$) 
& 0.000 & $5\times10^{-5}$ & $7.5\times10^{-5}$ & $1\times10^{-4}$ \\

\midrule
\multicolumn{5}{l}{\textit{Fixed variance guards}} \\

Shared variance guard ($\mathcal{L}_{\text{var}}^{\text{sh}}$) 
& 0.005 & 0.005 & 0.005 & 0.005 \\

Private variance floor ($\mathcal{L}_{\text{var}}^{\text{pr}}$) 
& 0.002 & 0.002 & 0.002 & 0.002 \\

\bottomrule
\end{tabular}

\vspace{2pt}
\footnotesize
The private standard-deviation floor was fixed to $0.20$ in all stages. 
Alignment used identity mode for epochs $<60$ and ConvAlign thereafter.
\end{table}

\begin{figure}[t]
  \centering
  \includegraphics[width=\textwidth]{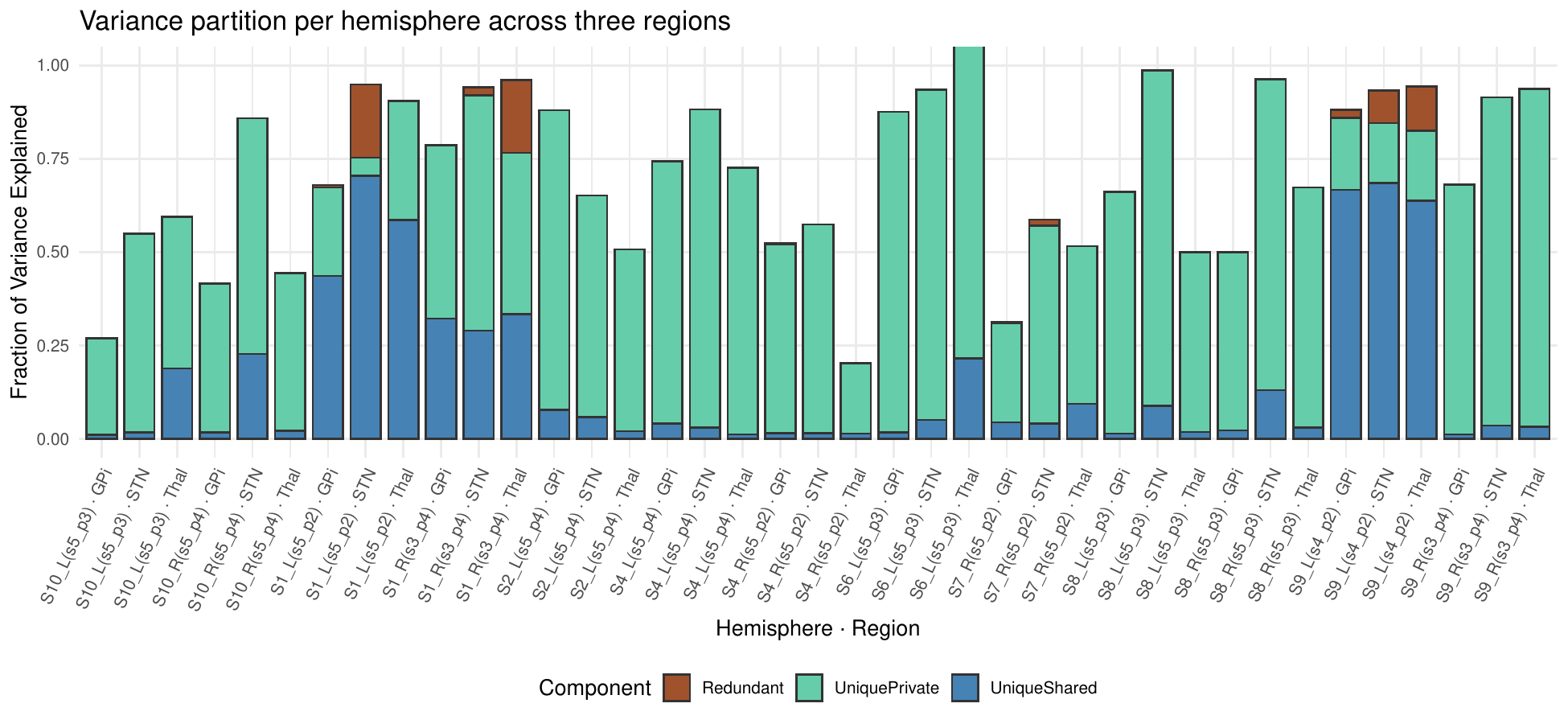}
  \caption{Subject-specific variance partition of GPi vs.\ STN vs.\ Thalamus
  For each subject-hemisphere, three adjacent bars (GPi, STN, Thal) show the fraction of variance explained by the \emph{unique shared}, \emph{unique private}, and \emph{redundant} components. Labels include the selected latent sizes $(d_s,d_p)$ per hemisphere chosen via the appendix rule~\ref{app:selection_rule}). Cohort heterogeneity is evident. Full dimension sweeps (per hemisphere and region) are in Appendix Fig.~\ref{fig:var_sweep3}.} 
  \label{fig:var_stack3}
\end{figure}

\begin{figure}[t]
  \centering

  \begin{subfigure}[b]{\textwidth}
    \centering
    \includegraphics[width=0.8\textwidth,clip=true,trim= 0in 0in 0in 0.3in]{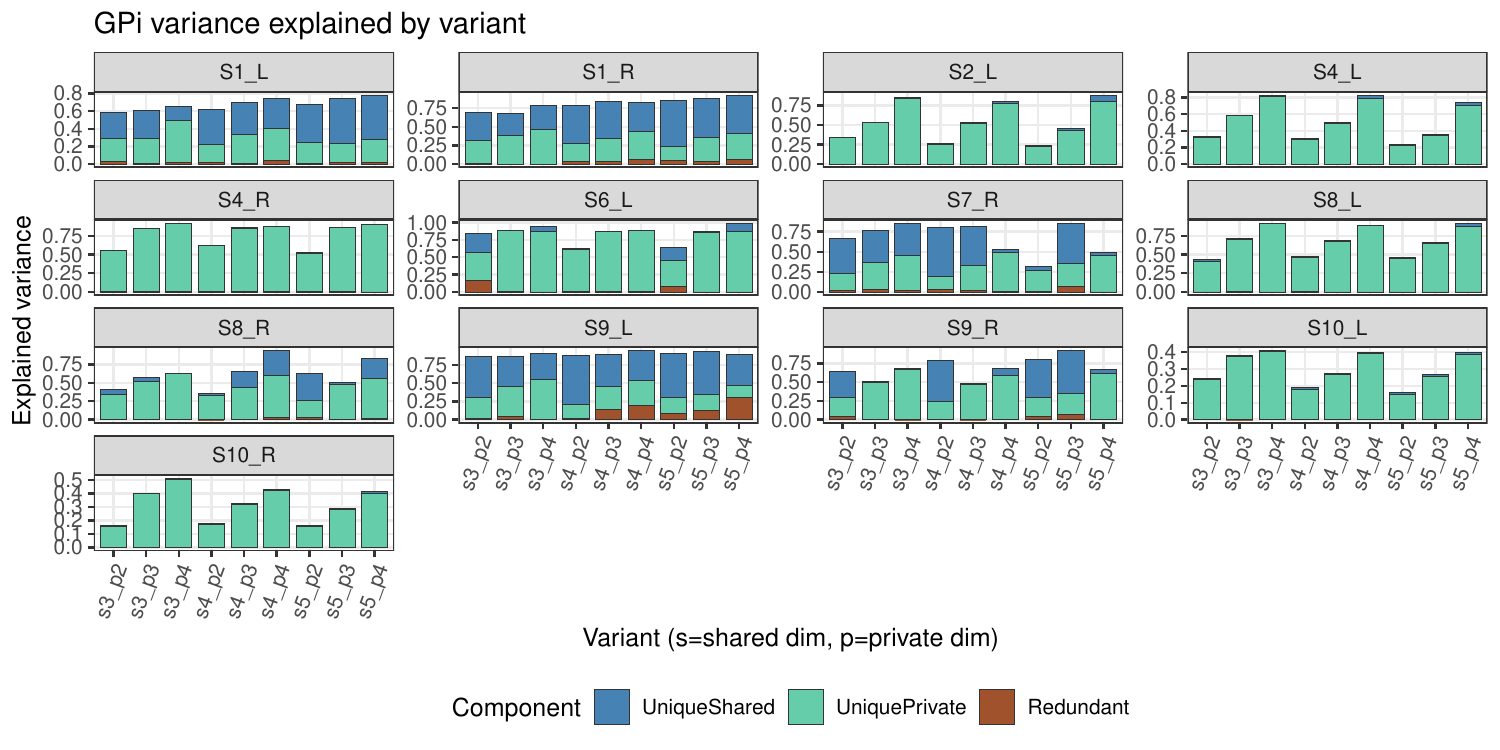}
    \caption{GPi}
    \label{fig:var_sweep_gpi3}
  \end{subfigure}

  \vspace{0.8em} 

  \begin{subfigure}[b]{\textwidth}
    \centering
    \includegraphics[width=0.8\textwidth,clip=true,trim= 0in 0in 0in 0.3in]{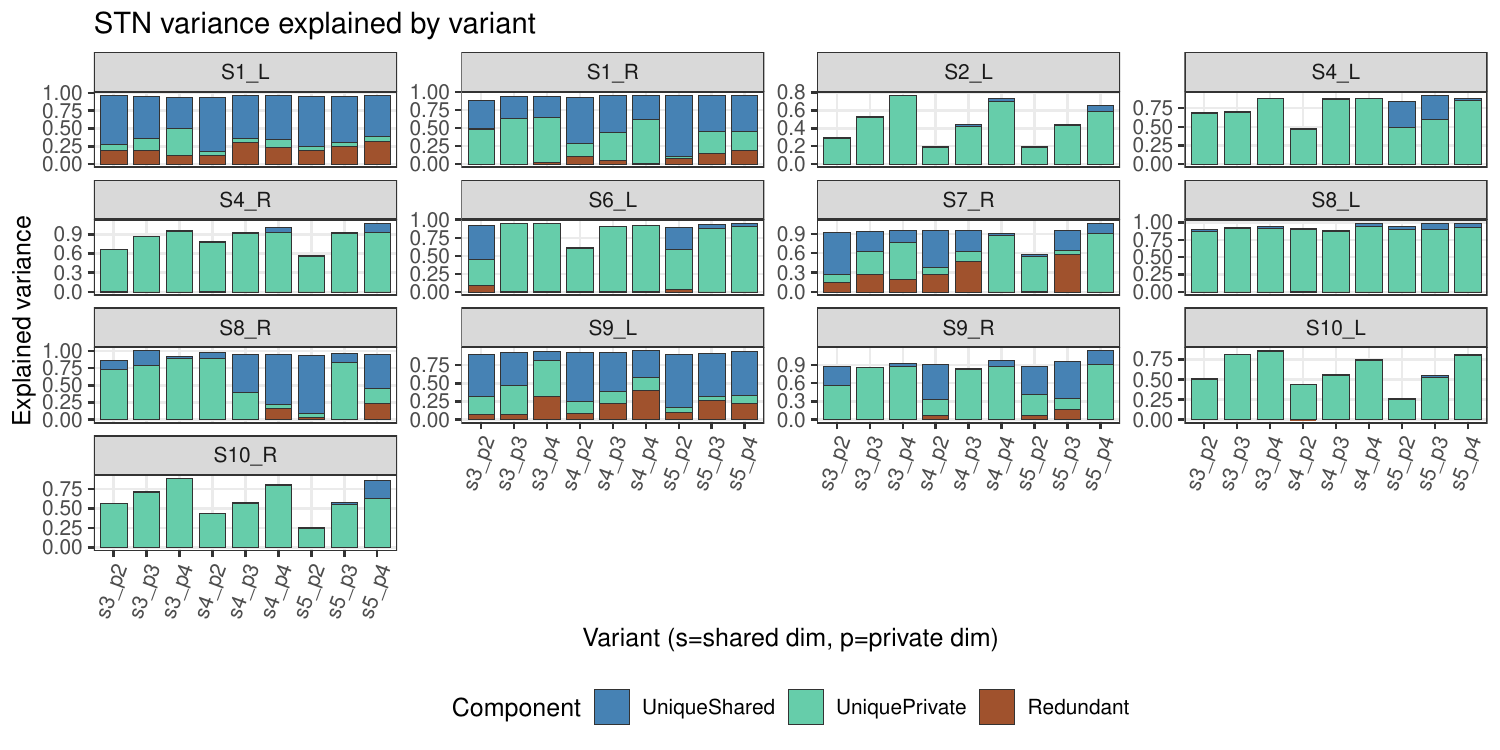}
    \caption{STN}
    \label{fig:var_sweep_stn3}
  \end{subfigure}

\vspace{0.8em} 
    
  \begin{subfigure}[b]{\textwidth}
    \centering
    \includegraphics[width=0.8\textwidth,clip=true,trim= 0in 0in 0in 0.3in]{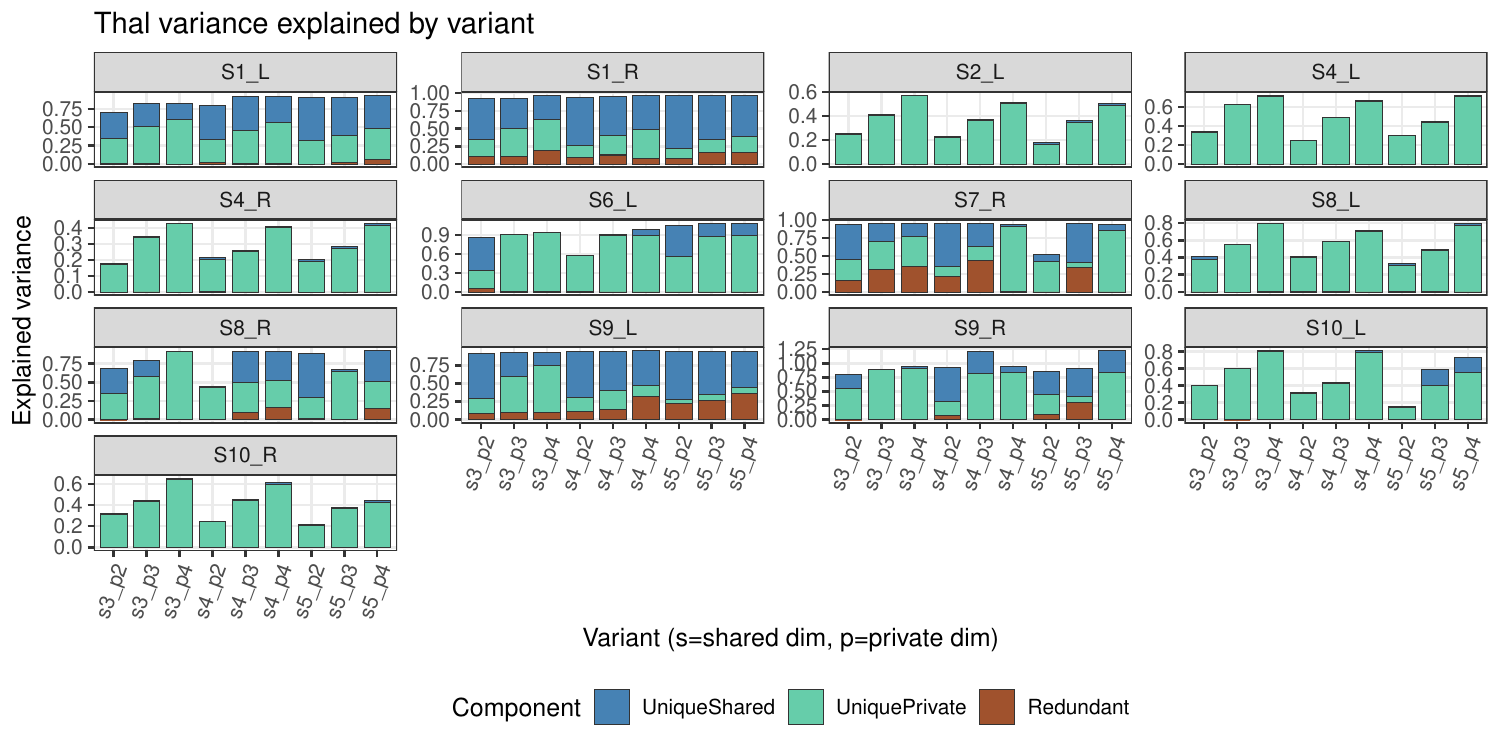}
    \caption{Thalamus}
    \label{fig:var_sweep_thal3}
  \end{subfigure}

  \caption{Variance partition across latent dimensionalities.
  For each subject–hemisphere, we trained SPIRE with shared dimension $d_s \in \{3,4,5\}$ and private dimension $d_p \in \{2,3,4\}$. Each bar shows the fraction of explained variance in the observed signal that is uniquely attributable to shared latents, uniquely attributable to private latents, or redundantly captured by both.  
  (a) GPi recordings. (b) STN recordings. (c) Thalamus recording. 
  These sweeps were used to select the final $(d_s,d_p)$ configuration per subject–hemisphere.}
  \label{fig:var_sweep3}
\end{figure}

\paragraph{Model selection.}
For each subject, we trained candidate models over private dimensions $d_{\mathrm{pr}}\in\{2,3,4\}$ and shared dimension $d_{\mathrm{sh}}\in\{3,4,5\}$. 
The final configuration was selected using the same criteria as the two-region analysis: low validation loss together with a non-degenerate variance partition between shared and private components. 
This ensured that adding thalamus did not collapse the representation into either a purely shared or purely private solution. 
The shared and private variance partitions for each selected subject-hemisphere are shown in Fig.~\ref{fig:var_stack3}, with the full dimensionality sweep shown in Fig.~\ref{fig:var_sweep3}.

\begin{figure}[t]
  \centering
  \includegraphics[width=\textwidth]{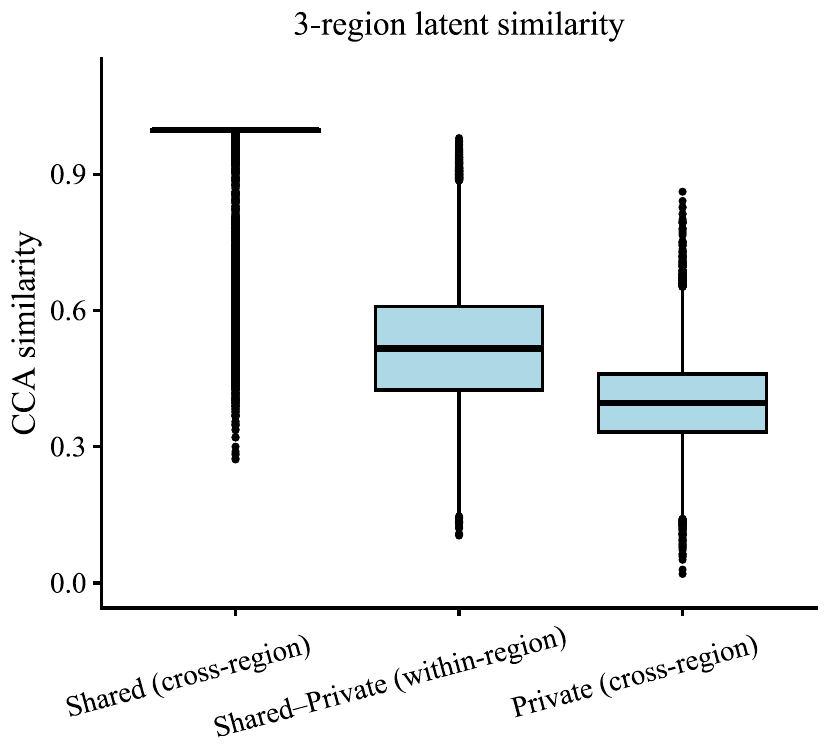}
  \caption{
Three-region latent similarity analysis. 
CCA similarity was computed among grouped latent pairs across GPi, STN, and thalamus. 
Shared cross-region latents showed the highest similarity, indicating recovery of a common multi-region latent manifold. 
Shared-private similarities were lower, reflecting limited within-region leakage, while private cross-region similarities were lowest, consistent with private latents capturing region-specific dynamics.
}
  \label{fig:CCA_3region}
\end{figure}

\paragraph{Latent alignment analysis}.
We also evaluated whether the three-region SPIRE model preserved meaningful shared/private structure by computing CCA similarity among latent subspaces. 
Shared latents aligned across different regions showed the highest similarity, indicating that SPIRE learned a common cross-regional manifold spanning GPi, STN, and thalamus. 
Shared-private similarities within the same region were lower, suggesting partial but limited leakage between shared and private components. 
Private-private similarities across regions were lowest, consistent with private latents capturing region-specific residual dynamics rather than shared network structure (Figure~\ref{fig:CCA_3region}).

\begin{figure}[t]
  \centering
  \includegraphics[width=\textwidth]{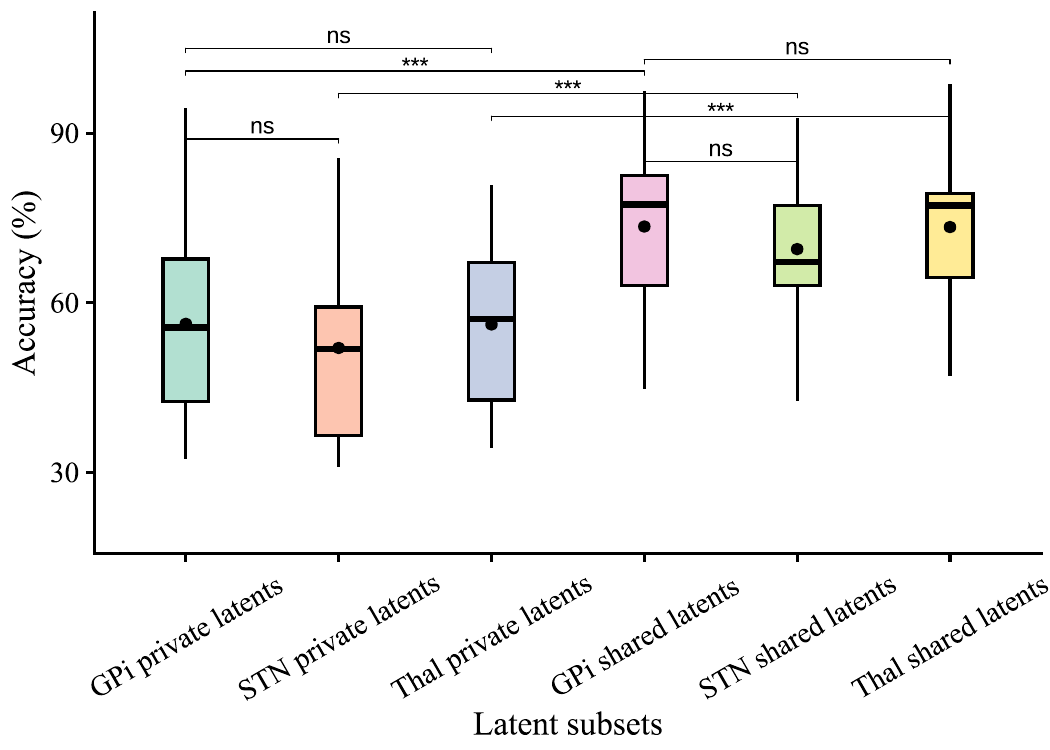}
\caption{
Three-region decoding of GPi stimulation frequency from SPIRE latents. 
Random Forest classifiers were trained to decode stimulation frequency (off, 85, 185, and 250 Hz) from private and shared latent subsets of GPi, STN, and thalamus. 
Shared latents across all three regions achieved higher decoding accuracy than private latents, indicating that GPi stimulation induces structured changes in the shared multi-region latent manifold. 
Private latents showed lower and more variable accuracy, consistent with region-specific residual activity. 
Boxplots show median and interquartile range, whiskers extend to 1.5$\times$IQR, and black dots denote means. 
Statistical comparisons are shown with conventional significance notation: ns, not significant; $^{***}p<0.001$.
}
  \label{fig:3region_classify}
\end{figure}

\paragraph{Three-region stimulation decoding.}
We next tested whether the three-region SPIRE model captured stimulation-dependent structure during GPi stimulation. 
Random Forest classifiers were trained to decode GPi stimulation frequency from each latent subset, using four conditions (off, 85, 185, and 250 Hz). 
Across GPi, STN, and thalamus, shared latents achieved higher decoding accuracy than private latents, with all shared representations performing well above chance. 
Notably, shared latents from STN and thalamus also encoded GPi stimulation frequency, indicating that stimulation effects were not restricted to the stimulated region but were expressed in the shared multi-region latent manifold. 
Private latents showed lower and more variable decoding accuracy, consistent with their role in capturing region-specific residual dynamics rather than distributed stimulation structure (Figure~\ref{fig:3region_classify}).

\subsubsection{Baseline models}
\label{app:baseline_training}

\textbf{CTAE}~\citep{sristi2025coupled} is a recently proposed transformer-based framework for learning coupled latent representations across multiple neural populations. 
CTAE employs shared and region-specific latent subspaces together with reconstruction, cross-reconstruction, alignment, and orthogonality objectives to model inter-regional neural structure. 
We selected CTAE as a strong nonlinear baseline due to its conceptual similarity to SPIRE, particularly its use of coupled latent spaces and explicit alignment objectives.

To adapt CTAE to intracranial LFP recordings from GPi and STN, we used the original two-region architecture without structural modifications. 
Following the original implementation, recordings from both regions were concatenated along the channel dimension and represented as multichannel temporal windows of shape \([N \times T \times C]\). 
All channels were z-scored prior to training. 
We trained CTAE using the same train/validation splits as SPIRE and matched latent dimensionalities to the corresponding SPIRE configurations for each subject. 
Models were trained using the Adam optimizer with learning rate \(10^{-4}\), two transformer layers, one attention head, fixed sinusoidal positional encodings, and early stopping based on validation reconstruction performance, and max 300 epochs.

CTAE produces time-resolved latent trajectories and therefore supports analyses requiring temporal latent dynamics, including latent similarity analysis and stimulation decoding. However, CTAE does not explicitly separate shared and private temporal dynamics through dedicated temporal alignment modules or variance-partitioning constraints. 

\textbf{Fundamental differences.} 
CTAE was developed for time-resolved shared/private neural representation
learning and is therefore the closest nonlinear baseline to SPIRE. It uses
transformer-based latent representations together with reconstruction,
cross-reconstruction, shared-latent alignment, and orthogonality objectives.
SPIRE targets the same general shared/private representation problem but differs
in several design choices motivated by continuous LFP: constrained temporal
ConvAlign operators, near-identity mapping regularization, shared/private
variance guards, shared-self reconstruction, and staged optimization of the
shared and private branches. These differences motivate our comparison under
nonlinear observation mappings, time-varying temporal offsets, and spectral
confounding, where stable assignment of transferable and region-specific
information is particularly challenging.

\textbf{Attempted DLAG baseline.}
We sought to benchmark SPIRE against DLAG~\citep{gokcen2022disentangling}, which has been successfully applied to multi-area spiking data. While DLAG worked on our synthetic datasets (Section~\ref{synth}), it proved infeasible on the human intracranial recordings. Specifically:
\begin{itemize}
    \item Ill-conditioned covariance matrices. After bipolar re-referencing and lag augmentation (Appendix~\ref{app:real_info}), the empirical covariance matrices often had deficient rank (e.g., eigenvalues $\approx 0$), causing the Gaussian process kernel inversion in DLAG to fail despite added jitter.
    \item Model mismatch. DLAG assumes stationary Gaussian processes with smoothly varying latents, whereas intracranial LFPs exhibited strong nonstationarities that violate these assumptions.
    \item Data dimensionality. Practical channel counts (3–8 bipolar contacts per region) combined with thousands of short windows led to numerical instabilities in the Toeplitz inversion routines central to DLAG. Even under aggressive dimensionality reduction or simplified hyperparameters, the model either diverged or collapsed to degenerate solutions.
\end{itemize}

We tried multiple remedies—reducing latent dimensionality, adding jitter, trimming channels, and using minimal configurations—but none yielded stable training. We therefore conclude that DLAG, in its current form, is not well suited to our bipolarized intracranial recordings. This limitation highlights the need for alternative frameworks, like SPIRE, that explicitly handle bipolar data, short windows, and nonstationary dynamics.

\paragraph{Compute comparison}

All experiments were conducted on a local workstation equipped with an NVIDIA RTX A4000 GPU (17.17~GB VRAM), Intel Xeon Silver 4216 CPU (32 logical processors), and 192~GB RAM. 

All measurements were performed in PyTorch using automatic mixed precision (AMP, fp16) and \texttt{torch.inference\_mode()}, with batch size \(B{=}8\) and inputs containing GPi/STN channels \((C_\mathrm{gpi}, C_\mathrm{stn}){=}(36,12)\). 
Each latency value represents the mean over 50 forward passes following 10 warm-up passes, with device synchronization before and after timing. 
SPIRE was evaluated using \texttt{private\_gate=1.0}. 
CTAE measurements used the original two-region transformer architecture with the same inference protocol as the real-data experiments.

We report two quantities per model (Table~\ref{tab:min-compute}): 
(i) trainable parameters (millions), and (ii) per-sequence inference latency (milliseconds). 
SPIRE was evaluated by a single forward pass on \((x_\mathrm{gpi}, x_\mathrm{stn})\). 
CTAE latency was measured using a single forward pass through the coupled transformer autoencoder without gradient computation.

\begin{table}[h]
\centering
\small
\begin{tabular}{lccc}
\toprule
Model & Params (M) & Seq length \(T\) & Inference (ms / seq) \\
\midrule
SPIRE (ours) & 0.116 & 247 & 4.45 \\
CTAE         & 0.860 & 247 & 4.84 \\
\bottomrule
\end{tabular}
\caption{
Compute comparison under matched batch size (\(B{=}8\)), channels \((36,12)\), and AMP (fp16). 
Latencies are wall-clock forward-pass times averaged over 50 runs following 10 warm-up iterations. 
}
\label{tab:min-compute}
\end{table}

\textbf{Note.} These measurements reflect inference only; training introduces additional backward-pass and optimizer overhead.

\textbf{DLAG:} Because DLAG is implemented in MATLAB and relies on an expectation–maximization (EM) procedure rather than a simple forward pass, we measured its compute cost separately. We used synthetic dataset D3 with 8 channels per region and prepared trial-wise sequences as in the DLAG demo code. To obtain an inference-like latency, we timed a single EM iteration (E–M step) with saving disabled using MATLAB’s \texttt{timeit}, then normalized by the number of sequences. Under these conditions, DLAG required on average $155.85$\,ms per sequence. We do not report parameter counts, since DLAG’s MATLAB implementation represents parameters via Gaussian process covariances and delay variables rather than explicit trainable tensors.

\subsection{Additional functional validation and robustness}
\label{app:robustness}
These analyses are secondary to the main narrative and evaluate robustness to lag augmentation, latent dimensionality, and alternative reconstruction objectives.

\subsubsection{Complete nonlinear ground-truth recovery matrix}
\label{app:nonlinear_matrix}
Using nonlinear Extra Trees regression with five-fold trial-wise cross-validation, we evaluated $SS$, $SP$, $PP$, and $PS$ as defined in Section~\ref{synth}. Values are mean (SD) across four matched seeds.
\begin{table}[h]
\caption{Complete nonlinear intended-versus-crossed recovery matrix. Specificity margins are $SS-SP$ and $PP-PS$.}
\label{tab:full_nonlinear_matrix}
\centering
\scriptsize
\begin{tabular}{llrrrrrr}
\toprule
Regime & Model & $SS$ & $SP$ & $PP$ & $PS$ & $SS-SP$ & $PP-PS$ \\
\midrule
D0 & SPIRE & 0.72 (.02) & 0.24 (.15) & 0.64 (.11) & 0.24 (.01) & 0.48 (.16) & 0.40 (.11) \\
   & DLAG  & 0.89 (.03) & 0.09 (.02) & 0.83 (.06) & 0.15 (.03) & 0.81 (.04) & 0.68 (.09) \\
   & CTAE  & 0.29 (.10) & 0.22 (.05) & 0.31 (.02) & 0.32 (.06) & 0.07 (.12) & -0.01 (.07) \\
\midrule
D1 & SPIRE & 0.76 (.01) & 0.31 (.03) & 0.60 (.01) & 0.20 (.01) & 0.44 (.02) & 0.39 (.02) \\
   & DLAG  & 0.64 (.01) & 0.21 (.01) & 0.48 (.03) & 0.28 (.01) & 0.43 (.01) & 0.20 (.02) \\
   & CTAE  & 0.14 (.05) & 0.12 (.02) & 0.17 (.04) & 0.18 (.04) & 0.02 (.05) & -0.01 (.02) \\
\midrule
D2 & SPIRE & 0.75 (.00) & 0.33 (.01) & 0.58 (.01) & 0.20 (.02) & 0.42 (.01) & 0.38 (.02) \\
   & DLAG  & 0.57 (.01) & 0.20 (.00) & 0.48 (.00) & 0.33 (.02) & 0.37 (.01) & 0.16 (.02) \\
   & CTAE  & 0.12 (.06) & 0.08 (.04) & 0.16 (.05) & 0.17 (.04) & 0.04 (.03) & -0.01 (.03) \\
\midrule
D3 & SPIRE & 0.54 (.02) & 0.11 (.02) & 0.55 (.04) & 0.28 (.01) & 0.42 (.00) & 0.27 (.03) \\
   & DLAG  & 0.42 (.03) & 0.26 (.02) & 0.46 (.01) & 0.33 (.02) & 0.15 (.04) & 0.13 (.02) \\
   & CTAE  & 0.17 (.05) & 0.23 (.04) & 0.35 (.03) & 0.22 (.04) & -0.06 (.02) & 0.14 (.05) \\
\bottomrule
\end{tabular}
\label{tab:CompletenonLin}
\end{table}
SPIRE satisfies $SS>SP$ and $PP>PS$ in every regime. DLAG shows the strongest factor segregation in linear D0, while its specificity degrades under nonlinear mappings, time-varying delay, and spectral confounding. CTAE shows approximately zero or negative specificity in several settings Table~\ref{tab:CompletenonLin}.

\subsubsection{No-lag retraining}
All real-data baselines in the original comparison received the same 0-3 sample lag-augmented inputs. To test whether lag stacking itself generates the reported organization, we retrained SPIRE without lag augmentation while keeping architecture, objective, schedule, optimizer, training duration, and stimulation-independent selection criteria fixed. Valid decompositions were obtained in 16/17 no-lag hemispheres versus 17/17 with lagging. Across 16 matched hemispheres, a hemisphere-level model showed a strong latent-pair effect ($F_{3,120}=33.80$, $p=6.59\times10^{-16}$), but no augmentation effect ($p=0.624$) and no augmentation-by-pair interaction ($p=0.961$).
\begin{table}[h]
\caption{Median CCA organization with and without lag augmentation across 16 matched hemispheres.}
\centering
\small
\begin{tabular}{lcc}
\toprule
Latent pair & Lag & No lag \\
\midrule
Shared GPi-STN & 0.987 & 0.750 \\
Shared-private GPi & 0.610 & 0.571 \\
Shared-private STN & 0.568 & 0.626 \\
Private GPi-STN & 0.390 & 0.406 \\
\bottomrule
\end{tabular}
\end{table}
The reconstruction hierarchy and stimulation conclusion were preserved. Under GPi stimulation, all four shared-versus-private centroid-shift contrasts remained significant without lagging ($p<0.0001$). Under STN stimulation, all four retained the same direction and three remained significant after correction. Lag stacking therefore improves fitting robustness but is not required to produce the main shared-over-private stimulation result.

\subsubsection{Latent-dimensionality sensitivity}
We evaluated all nine combinations $d_{\mathrm{sh}}\in\{3,4,5\}$ and $d_{\mathrm{pr}}\in\{2,3,4\}$ using configurations selected only by off-stimulation variance criteria. Across 64 hemisphere-configuration models for GPi stimulation and 52 for STN stimulation, there was a significant overall shared-versus-private stimulation effect ($F=16.15$, $p=8.2\times10^{-5}$), but no interaction with shared dimension ($p=0.410$) or private dimension ($p=0.176$). The effect magnitude differed by stimulation site. A conservative analysis retaining partially collapsed configurations also preserved an overall shared/private effect with no dimension interaction. Thus, dimension dependence was not detected for the principal contrast.

\subsubsection{Private-only reconstruction objective}
We evaluated the performance adding another loss term requiring the private branch alone to reconstruct the complete same-region signal. At a relatively strong staged weight (maximum 0.03), only 2/17 hemispheres retained any configuration satisfying the predefined variance-partition criteria; most failures reflected excessive shared/private redundancy or insufficient unique shared variance. Under controlled synthetic ground truth, smaller maximum weights of 0.01 and 0.005 did not produce a uniform improvement in CCA or nonlinear intended recovery/specificity after correction Table~\ref{tab:CCAprivOnly}.
\begin{table}[h]
\caption{CCA shared/private specificity margins for the original objective and two lower-weight private-only reconstruction variants.}
\centering
\small
\begin{tabular}{lccc}
\toprule
Regime & Original & Private 0.01 & Private 0.005 \\
\midrule
D0 & 0.360 / 0.397 & 0.455 / 0.446 & 0.454 / 0.455 \\
D1 & 0.644 / 0.548 & 0.647 / 0.526 & 0.645 / 0.548 \\
D2 & 0.607 / 0.539 & 0.610 / 0.524 & 0.606 / 0.540 \\
D3 & 0.472 / 0.463 & 0.482 / 0.475 & 0.478 / 0.480 \\
\bottomrule
\end{tabular}
\label{tab:CCAprivOnly}
\end{table}
We additionally retrained all 17 human hemispheres with maximum $\lambda_{\mathrm{private}}=0.005$ while retaining each hemisphere's previously selected dimensionality. Private-only reconstruction improved (GPi MSE 0.8659$\rightarrow$0.482; STN 0.6110$\rightarrow$0.376), while full and shared reconstruction remained similar and cross-region private reconstruction remained poor. However, nonlinear organization exposed a tradeoff: excess $R^2$ was 0.574 for shared-shared, 0.530 for shared-to-private, 0.315 for private-to-shared, and -0.001 for private-private cross-region prediction. The shared-shared minus shared-to-private margin therefore fell from 0.132 in the original model to 0.044 in the modified model. The main stimulation findings remained unchanged. This variant is viable, but it does not uniformly improve functional separation, motivating retention of the original objective.

\subsection{Secondary human-LFP diagnostics}
\label{app:human_secondary}
\begin{figure}[h]
    \centering
    \begin{subfigure}[t]{0.95\linewidth}
        \centering
        \includegraphics[width=\linewidth]{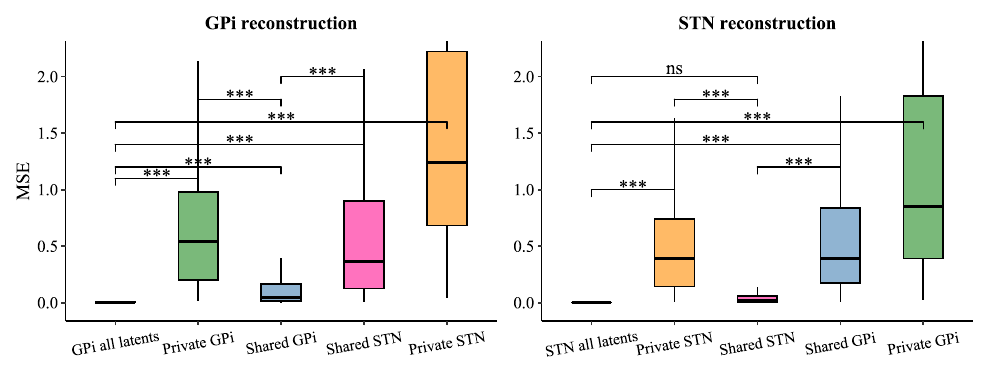}
        \caption{Two-region reconstruction.}
    \end{subfigure}
    \vspace{0.4em}
   
    \hfill
    \begin{subfigure}[t]{0.95\linewidth}
        \centering
        \includegraphics[width=\linewidth]{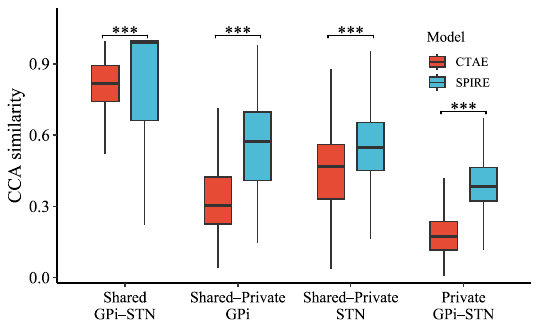}
        \caption{CCA organization and CTAE comparison.}
    \end{subfigure}
    \caption{
Secondary held-out human-LFP diagnostics.
(a) Reconstruction error when decoding GPi and STN activity from full,
shared-only, and private-only representations.
(b) CCA similarity between shared and private latent spaces for SPIRE and CTAE.
These analyses provide complementary characterization of representation
quality and organization; the primary evidence for functional segregation is
the nonlinear cross-region prediction analysis in the main text.
}
    \label{fig:human_secondary}
\end{figure}

The main text uses nonlinear cross-region predictability as the primary
functional validation of the shared/private decomposition. Here we provide
additional held-out diagnostics that characterize complementary properties of
the learned representation, including reconstruction quality, latent
organization, and similarity between SPIRE and CTAE latent spaces Figure~\ref{fig:human_secondary}. These
analyses are not used to define the shared/private interpretation but provide
additional characterization of the learned representations.

\FloatBarrier

\subsection{Frozen-representation validation across Train-Off, Test-Off, and Test-On}
\label{app:train_test_on}
This analysis verifies that models learned and selected on baseline data remain usable on stimulation recordings; it is not presented as a separate contribution. No model was retrained, fine-tuned, or reselected using stimulation data. We compared sampled baseline training windows (Train-Off), held-out baseline recordings (Test-Off), and held-out stimulation recordings (Test-On) in 17 GPi-stimulation and 13 STN-stimulation hemispheres.

Full reconstruction $R^2$ decreased under stimulation as expected for a distribution shift, but remained high Table~\ref{tab:reconOn}.
\begin{table}[h]
\caption{Mean Test-On full-reconstruction $R^2$.}
\centering
\small
\begin{tabular}{lcc}
\toprule
Cohort & GPi target & STN target \\
\midrule
GPi stimulation & 0.888 & 0.952 \\
STN stimulation & 0.853 & 0.905 \\
\bottomrule
\end{tabular}
\label{tab:reconOn}
\end{table}
The objective-related hierarchy full error $<$ shared-self error $<$ aligned cross-shared error was preserved for both regions in every hemisphere (17/17 under GPi stimulation and 13/13 under STN stimulation), and aligned cross-shared reconstruction did not significantly degrade from Test-Off to Test-On. Latent variance showed no significant overall stage effect ($p=0.282$ for the GPi-stimulation cohort; $p=0.201$ for STN stimulation), and shared SD changed by only -1.8\% to +3.2\% from Test-Off to Test-On. Shared-private dependence increased modestly under stimulation but no region-specific Test-On versus Test-Off contrast survived correction.

Pointwise shared-alignment error increased under stimulation (48\% for GPi stimulation and 60\% for STN stimulation), which is compatible with the intended use of the representation to measure reorganization. As a secondary check not used during training, shared GPi-STN CCA remained larger than private GPi-STN CCA in every hemisphere~\ref{tab:CCAon}.
\begin{table}[h]
\caption{Functional organization of the frozen representation on unseen stimulation data.}
\centering
\small
\begin{tabular}{lccc}
\toprule
Cohort & Shared CCA & Private CCA & Shared $>$ private \\
\midrule
GPi stimulation & 0.588 & 0.189 & 17/17 \\
STN stimulation & 0.516 & 0.197 & 13/13 \\
\bottomrule
\end{tabular}
\label{tab:CCAon}
\end{table}
Thus stimulation changes geometry within the learned coordinate system without causing collapse or reversal of the defining shared/private organization.

\subsection{Matched CTAE stimulation details}
\label{app:CTAE_onstim_additional}
The main text reports the class-averaged CTAE-SPIRE comparison. Region-specific CTAE mean accuracies were 31.0\% for private GPi, 31.0\% for private STN, 38.2\% for shared GPi, and 38.5\% for shared STN. Averaged across regions, CTAE shared accuracy exceeded private accuracy by 7.41 points ($F_{1,48}=44.53$, $p=2.34\times10^{-8}$), with no region effect ($p=0.908$) and no class-by-region interaction ($p=0.893$). SPIRE exceeded CTAE separately for private GPi, shared GPi, private STN, and shared STN; both class-averaged differences in Table~\ref{tab:ctae_stim} had Holm-adjusted $p<0.0001$.

\subsection{External CA1-PFC LFP transfer example}
\label{app:external_rat}
To evaluate applicability beyond the pediatric GPi-STN dataset, we applied the complete SPIRE architecture/objective to a distinct public rat CA1-PFC LFP dataset. This is a preliminary transfer example, not a main claim of universal or tuning-free generalization. The numerical optimization schedule was adapted once at the dataset level, then held fixed across all four animals. Every rat used three shared and three private dimensions with no rat-specific hyperparameter or dimension search. Models were trained separately using complete epochs split 70/15/15 into training, validation, and untouched test sets.

\begin{table}[h]
\caption{Held-out CA1-PFC latent organization in four rats.}
\centering
\small
\begin{tabular}{lccc}
\toprule
Rat & Shared CCA & Private CCA & Within-region sh-pr CCA \\
\midrule
T1 & 0.987 & 0.064 & 0.135-0.190 \\
S2 & 0.995 & 0.022 & 0.038-0.122 \\
Q1 & 0.997 & 0.067 & 0.033-0.052 \\
R1 & 0.660 & 0.049 & 0.213-0.230 \\
\bottomrule
\end{tabular}
\label{tab:CCArat}
\end{table}
The intended ordering held in 4/4 animals; mean shared cross-region CCA was 0.910 versus 0.050 for private cross-region CCA~\ref{tab:CCArat}.

\begin{table}[h]
\caption{Nonlinear held-out cross-predictability in the CA1-PFC transfer example.}
\centering
\small
\begin{tabular}{lccc}
\toprule
Rat & Shared-shared $R^2$ & Shared-to-private $R^2$ & Private-private $R^2$ \\
\midrule
T1 & 0.978 & 0.323 & -0.008 \\
S2 & 0.990 & 0.461 & -0.018 \\
Q1 & 0.997 & 0.345 & -0.013 \\
R1 & 0.620 & 0.154 & -0.016 \\
\bottomrule
\end{tabular}
\label{tab:R2rat}
\end{table}
Cross-region private prediction remained at the structured temporal null while shared prediction remained strong in every animal~\ref{tab:R2rat}.

\begin{table}[h]
\caption{Held-out reconstruction MSE in the external LFP transfer example; each cell is full/shared-only/private-only.}
\centering
\small
\begin{tabular}{lcc}
\toprule
Rat & CA1 & PFC \\
\midrule
T1 & 0.0044 / 0.158 / 0.863 & 0.0358 / 0.182 / 1.120 \\
S2 & 0.0008 / 0.036 / 0.276 & 0.0245 / 0.267 / 1.074 \\
Q1 & 0.0017 / 0.059 / 0.686 & 0.0260 / 0.196 / 2.566 \\
R1 & 0.0025 / 0.175 / 0.809 & 0.0361 / 0.171 / 0.965 \\
\bottomrule
\end{tabular}
\label{tab:reconrat}
\end{table}
The reconstruction hierarchy full $<$ shared-only $<$ private-only held for both regions in all four rats. Four concordant animals give a one-sided exact sign-test value of $p=0.0625$, so we emphasize consistency rather than population-level inference~\ref{tab:reconrat}. This experiment shows practical dataset-level adaptation without animal-specific redesign; it does not establish universal transfer.
\FloatBarrier

\subsection{Limitations of larger multi-region and multi-subject extensions}
\label{app:extension_multi}
For $R$ regions, SPIRE instantiates $R(R-1)$ directed alignment pathways. Region-wise losses are averaged across regions and pairwise losses across directed pairs, so the number of coefficient \emph{types} does not increase, but memory and computation scale quadratically in the dense implementation. Sparse anatomical/data-driven alignment graphs or alignment to a learned reference space are natural extensions.

The present one-shared/one-private formulation also does not explicitly distinguish globally shared factors from factors shared only within a subset of regions. A subnetwork-specific process may be represented partly in shared space, partly in private space, or weakened by alignment to less-related regions. The current three-region analysis demonstrates recovery of a common GPi-STN-thalamic manifold, not arbitrary subnetwork discovery. Hierarchical global/subnetwork/private latents remain future work.

Finally, current models are fit separately by hemisphere. A multi-subject extension could combine subject-specific input projections/decoders with a common region-level shared space and subject-specific private components, informed by anatomical labels or electrode coordinates. Because electrodes sample different tissue across subjects, cross-subject correspondence cannot be assumed and is not established here.

\end{document}